\documentclass[review,10pt]{JMtemplate}

\usepackage{times}
\usepackage{graphicx}
\usepackage{bm}
\usepackage{paralist,algorithmic,algorithm}
\usepackage{amsmath,mathrsfs,amssymb}
\usepackage{url}            
\usepackage{amsfonts}       
\usepackage{nicefrac}       
\usepackage{microtype}      
\usepackage{booktabs}       
\usepackage{longtable}
\usepackage{threeparttable}
\usepackage{multirow}
\usepackage{soul}
\usepackage{subfigure}
\usepackage{wrapfig}
\usepackage{tikz}
\usetikzlibrary{fpu}
\usetikzlibrary{math}
\usepackage[american]{babel}
\usepackage{setspace}
\usepackage{stfloats}
\usepackage{lipsum}
\usepackage{multicol}
\usepackage{pifont}  
\usepackage{makecell}
\usepackage{algorithm}
\usepackage{algorithmic}
\usepackage{hyperref}

\usepackage{natbib}  
\usepackage{caption} 

\newtheorem{theorem}{Theorem}
\newtheorem{definition}{Definition}
\newtheorem{corollary}{Corollary}

\newcommand*{\dif}{\mathop{}\!\mathrm{d}}
\newcommand*{\tr}{\mathop{}\!\mathrm{tr}}

\newcommand*{\e}{\mathop{}\!\mathrm{e}}

\begin{document}
\begin{frontmatter}
\title{A Unified Kernel for Neural Network Learning}

\author{Shao-Qun Zhang\textsuperscript{\rm 1,2,}\footnote{Shao-Qun Zhang is the corresponding author.}}
\author{Zong-Yi Chen\textsuperscript{\rm 1,2}}
\author{Yong-Ming Tian\textsuperscript{\rm 1,2}}
\author{Xun Lu\textsuperscript{\rm 1,2}}

\address{\textsuperscript{\rm 1}State Key Laboratory of Novel Software Technology, Nanjing University, China \\
	\textsuperscript{\rm 2}School of Intelligent Science and Technology, Nanjing University, China \\
    \{zhangsq, chenzy, tianym\}@lamda.nju.edu.cn, lux2022@smail.nju.edu.cn}
\date{\today}

\begin{abstract}
Past decades have witnessed a great interest in the distinction and connection between neural network learning and kernel learning. Recent advancements have made theoretical progress in connecting infinite-wide neural networks and Gaussian processes. Two predominant approaches have emerged: the Neural Network Gaussian Process (NNGP) and the Neural Tangent Kernel (NTK). The former, rooted in Bayesian inference, represents a zero-order kernel, while the latter, grounded in the tangent space of gradient descents, is a first-order kernel. In this paper, we present the Unified Neural Kernel (UNK), which  {is induced by the inner product of produced variables and characterizes the learning dynamics of neural networks with gradient descents and parameter initialization.} The proposed UNK kernel maintains the limiting properties of both NNGP and NTK, exhibiting behaviors akin to NTK with a finite learning step and converging to NNGP as the learning step approaches infinity. Besides, we also theoretically characterize the uniform tightness and learning convergence of the UNK kernel, providing comprehensive insights into this unified kernel. Experimental results underscore the effectiveness of our proposed method.
\end{abstract}

\begin{keyword}
Neural Network Learning \sep
Unified Neural Kernel \sep
Neural Network Gaussian Process \sep
Neural Tangent Kernel \sep
Gradient Descent \sep
Parameter Initialization \sep
Uniform Tightness \sep
Convergence
\end{keyword}
\end{frontmatter}

\section{Introduction}  \label{sec:introduction}
While neural network learning is successful in a number of applications, it is not yet well understood theoretically \citep{poggio2020theoretical}. Recently, there has been an increasing amount of literature exploring the correspondence between infinite-wide neural networks and Gaussian processes~\citep{neal1996:GP}. Researchers have identified equivalence between the two in various architectures~\citep{garriga2019:GP,novak2018:GP,yang2019:GP}. This equivalence facilitates precise approximations of the behavior of infinite-wide Bayesian neural networks without resorting to variational inference. Relatively, it also allows for the characterization of the distribution of randomly initialized neural networks optimized by gradient descent, eliminating the need to actually run an optimizer for such analyses.

The standard investigation in this field encompasses the Neural Network Gaussian Process (NNGP)~\citep{lee2018:NNGP}, which establishes that a neural network converges to a Gaussian process statistically as its width approaches infinity. The NNGP kernel inherently induces a posterior distribution that aligns with the feed-forward inference of infinite-wide Bayesian neural networks employing an i.i.d. Gaussian prior. Another typical work is the Neural Tangent Kernel (NTK)~\citep{jacot2018:NTK}, where the function of a neural network trained through gradient descent converges to the kernel gradient of the functional cost as the width of the neural network tends to infinity. The NTK kernel captures the learning dynamic wherein learned parameters are closely tied to their initialization, resembling an i.i.d. Gaussian prior. These two kernels, derived from neural networks, exhibit distinct characteristics based on different initializations and regularization. A notable contrast lies in the fact that the NNGP, rooted in Bayesian inference, represents a zero-order kernel that is more suitable to describe the overall characteristics of neural network learning. In contrast, the NTK, rooted in the tangent space of descent gradients, is a first-order kernel that is adept at capturing local characteristics of neural network learning. Empirical evidence provided by Lee et al.~\citep{lee2020finite} demonstrates the divergent generalization performances of these two kernels across various datasets.
\begin{table}[!htb]
	\centering
	\caption{Key properties of our UNK kernel compared to the NNGP and NTK kernels, where $\star\star$ denotes no results so far.}
	\label{tab:theory}
		\begin{tabular}{l | c | c | c}
			\toprule
			\textbf{Properties}   & \textbf{NNGP} & \textbf{NTK} & \textbf{UNK}  \\
			\midrule
			Existing & \makecell{\citep[Section 2]{lee2018:NNGP},\\ \citep[Theorem 1]{zhang2022:nngp}} & \citep[Theorem 1]{jacot2018:NTK} & Theorem~\ref{thm:unified} \\ \midrule
			Limiting Properties & when $\lambda = 0$ or $t=0$ & when $\lambda \neq 0$ and $t \to \infty$ &  Theorem~\ref{thm:unified} \\ \midrule
			Uniform Tightness &  \makecell{\citep[Lemma 3]{bracale2020:asymptotic},\\ \citep[Theorem 2]{zhang2022:nngp}}  & $\star\star$ & Theorem~\ref{thm:asymptotic} \\ \midrule
			Convergence & \citep[Theorem 3]{zhang2022:nngp}  & \makecell{\citep[Theorem 2]{jacot2018:NTK},\\ \citep[Theorem 3.2]{nguyen2021:eigenvalues}} & Theorem~\ref{thm:smallest} \\ 
			\bottomrule
	\end{tabular} 
\end{table}

In this paper, we endeavor to unify the NNGP and NTK kernels and present the Unified Neural Kernel (UNK) as a cohesive framework for neural network learning. The proposed UNK kernel is induced by the inner product of produced variables and built upon the learning dynamics associated with gradient descents and parameter initialization and unifies the limiting properties of both the NTK and NNGP kernels. This work delves into theoretical characterizations, including but not limited to the existence, limiting properties, uniform tightness, and learning convergence of the proposed UNK kernel. Table~\ref{tab:theory} shows our progress compared to the NNGP and NTK kernels. Our theoretical investigations reveal that the UNK kernel exhibits behaviors reminiscent of the NTK kernel with a finite learning step and converges to the NNGP kernel as the learning step approaches infinity. We also conduct experiments on benchmark datasets using various configurations. The numerical results further underscore the effectiveness of our proposed UNK kernel. These results significantly expand the scope of the existing theory connecting kernel learning and neural network learning.

The rest of this paper is organized as follows. Section~\ref{sec:pre} introduces useful notations and related studies. Section~\ref{sec:unify} presents the UNK kernel with in-depth discussions and proof sketches. Section~\ref{sec:properties} shows the uniform tightness and convergence of the UNK kernel. Section~\ref{sec:experiments} conducts numerical experiments. Section~\ref{sec:conclusions} concludes our work.

\section{Preliminary}  \label{sec:pre}
We start preliminaries with useful notations. Let $[N] = \{1, 2, \dots, N\}$ be an integer set for $N \in \mathbb{N}^+$, and $|\cdot|_{\#}$ denotes the number of elements in a collection, e.g., $|[N]|_{\#} = N$. We define the globe $\mathcal{B}(r) = \{ \boldsymbol{x} \mid \| \boldsymbol{x} \|_2 \leq r \}$ for any $r\in\mathbb{R}^+$. Let $\mathbf{I}_n$ be the $n \times n$-dimensional identity matrix. Let $\|\cdot\|_p$ be the norm of a vector or matrix, in which we employ $p=2$ as the default. Given $\boldsymbol{x}=(x_1,\dots,x_n)$ and $\boldsymbol{y}=(y_1,\dots,y_n)$, we also define the sup-related measure as $\| \boldsymbol{x} - \boldsymbol{y} \|_{\alpha}^{\textrm{sup}} = \sup_{i\in[n]} \big| x_i - y_i \big|^{\alpha}$ for $\alpha>0$. Let $\mathcal{C}(\mathbb{R}^{n_0};\mathbb{R}^n)$ be the space of continuous functions where $n_0,n\in\mathbb{N}$. Provided a linear and bounded functional $\mathcal{F}: \mathcal{C}(\mathbb{R}^{n_0};\mathbb{R}^n) \to \mathbb{R}$ and a function $f \in  \mathcal{C}(\mathbb{R}^{n_0};\mathbb{R}^n)$ which satisfies $f(\boldsymbol{x}) \overset{\underset{\mathrm{d}}{}}{\to} f^*$, then we have $\mathcal{F} (f(\boldsymbol{x})) \overset{\underset{\mathrm{d}}{}}{\to} \mathcal{F}(f^*)$ and $\mathbb{E} \left[ \mathcal{F} (f(\boldsymbol{x})) \right] \to \mathbb{E} \left[ \mathcal{F}(f^*) \right]$ according to General Transformation Theorem~\citep[Theorem 2.3]{van2000asymptotic} and Uniform Integrability~\citep{billingsley2013convergence}, respectively.

Throughout this paper, we use the specific symbol $K$ to denote the concerned kernel for neural network learning. The superscript $(l)$ and stamp $t$ are used for recording the indexes of hidden layers and training epochs, respectively. We denote the Gaussian distribution by $\mathcal{N}(\mu_x, \sigma_x^2)$, where $\mu_x$ and $\sigma_x^2$ indicate the mean and variance, respectively. In general, we employ $\mathbb{E}(\cdot)$ and $\mathrm{Var}(\cdot)$ to denote the expectation and variance, respectively. To leverage the convergence rate, we also introduce the limiting complexity. Given two functions $g,h\colon \mathbb{N}^+\rightarrow \mathbb{R}$ relative to $n$, we denote by $h=\mathbf{\Theta}(g)$ if there exist positive constants $c_1,c_2$, and $n_0$ such that $c_1g(n) \leq h(n) \leq c_2g(n)$ for every $n \geq n_0$; $h=\mathcal{O}(g)$ if there exist positive constants $c$ and $n_0$ such that $h(n) \leq cg(n)$ for every $n \geq n_0$; $h=\Omega(g)$ if there exist positive constants $c$ and $n_0$ such that $h(n) \geq cg(n)$ for every $n \geq n_0$.

\subsection{NNGP and NTK}  \label{subsec:two_kernels}
We start this work with an $L$-hidden-layer fully-connected neural network,
\begin{equation} \label{eq:feedforward}
	\left\{~\begin{aligned}
		\boldsymbol{h}^{(l)} &= \mathbf{W}^{(l)} \boldsymbol{s}^{(l-1)} + \boldsymbol{b}^{(l)} \ ,\quad \text{for}\quad l \in [L] \ , \\
		\boldsymbol{s}^{(0)} &= \boldsymbol{x} \quad\text{and}\quad  \boldsymbol{s}^{(l)} = \phi( \boldsymbol{h}^{(l)} ) \ ,\quad\text{for}\quad l \in [L] \ ,\\
		\boldsymbol{y} &= \boldsymbol{s}^{L} \ ,
	\end{aligned}\right.
\end{equation}
in which $\boldsymbol{x} \in \mathbb{R}^{n_0}$ and $\boldsymbol{y} \in \mathbb{R}^{n_L}$ indicate the variables of inputs respectively, $\boldsymbol{h}^{(l)} \in \mathbb{R}^{n_l}$ and $\boldsymbol{s}^{(l)} \in \mathbb{R}^{n_l}$ denote the pre-synaptic and post-synaptic variables of the $l$-th hidden layer respectively,  $n_l$ and $n_0$ indicate the number of neurons in the $l$-th hidden layer for $l \in [L]$ and the input layer respectively, $\mathbf{W}^{(l)} \in \mathbb{R}^{n_l \times n_{l-1}} $ and $\boldsymbol{b}^{(l)} \in \mathbb{R}^{n_l} $ are the parameter variables of connection weights and bias respectively, and $\phi$ is an element-wise activation function.  {We here name $\boldsymbol{s}^{(l)}$ the \emph{produced variables}.} For convenience, we here note the parameter variables at the $t$-th epoch as $\Theta^{(l)}_t = [ \mathbf{W}^{(l)}, \boldsymbol{b}^{(l)} ]$, and $\Theta^{(l)}_0$ denotes the initialized parameters, which obeys the Gaussian distribution $\mathcal{N}(0, \sigma^2/n_l)$.

\paragraph{\bf Neural Network Gaussian Process (NNGP)} For any $l \in [L]$, there is a claim that the conditional variable $\boldsymbol{h}^{(l)} |\boldsymbol{s}^{(l-1)}$ obeys the Gaussian distribution.  {In detail, $\boldsymbol{h}^{(l)}$ is a sum of i.i.d. random terms according to
	\[
	\boldsymbol{h}^{(l)} = \mathbf{W}^{(l)} \boldsymbol{s}^{(l-1)} + \boldsymbol{b}^{(l)}
	\quad\text{with}\quad
	\boldsymbol{s}^{(l-1)}  = \phi \left( \boldsymbol{h}^{(l-1)} \right) \ .
	\]
	Thus, as $n_l \to \infty$, $\boldsymbol{h}^{(l)}$ will maintain joint multivariate Gaussian distribution, i.e., 
	\[
	\boldsymbol{h}^{(l)} |	\boldsymbol{s}^{(l-1)}  \sim \mathcal{N} (  \boldsymbol{0},  \eta^{\textrm{NNGP}}_{l-1} \sigma^2 \mathbf{I}_{n_l} )
	\]
	or equally, the inner product of produced variables induced a Gaussian distribution
	\[
	\lim\limits_{n_{l-1} \to \infty} \mathbb{E} \left\langle \boldsymbol{h}^{(l)} |\boldsymbol{s}^{(l-1)}, \boldsymbol{h}^{(l)} | \boldsymbol{s}^{(l-1)}  \right\rangle = \mathbf{I}_{n_l} \eta^{\textrm{NNGP}}_{l-1} \sigma^2 \ ,
	\]
	where $\eta^{\textrm{NNGP}}_{l-1} = (1+C_{\phi}) /(n_{l-1} C_{\phi}) $ and $C_{\phi} = {1}/{\mathbb{E}_{z \sim \mathcal{N}(0,1)} \left( \phi(z) \right)^2 }$.} Further, the NNGP kernel is defined by 
\[
K_{\textrm{NNGP}}^{(l)} \left( \boldsymbol{s}^{(l-1)}, \boldsymbol{s}'^{(l-1)} \right) = \sigma^2 \mathbb{E} \langle\boldsymbol{s}^{(l-1)},  \boldsymbol{s}'^{(l-1)} \rangle + \sigma^2 \ .
\]
It is observed that the NNGP is closely related to the initial parameters.

\paragraph{\bf Neural Tangent Kernel (NTK)} The training of the concerned neural network consists in optimizing $\boldsymbol{y} = f(\boldsymbol{x} ; \Theta)$ in the function space, supervised by a functional loss $\mathcal{L}(\Theta)$, such as the square or cross-entropy functions
\begin{equation}  \label{eq:gd}
	\frac{\dif \Theta}{\dif t} = - \frac{\dif \mathcal{L}(\Theta)}{\dif \Theta} = - \frac{\dif \mathcal{L}(\Theta)}{\dif f(\boldsymbol{x} ; \Theta)} \frac{\dif f(\boldsymbol{x} ; \Theta)}{\dif \Theta} \ ,
\end{equation}
where $\Theta$ denotes the variable of any parameter. For any $l \geq 2$,  {there is a claim that the inner product of gradient variable induces a Gaussian distribution. Taking $\mathbf{W}^{(l-1)}$ as an example, we have 
	\[
	\left\langle \frac{\partial \boldsymbol{h}^{(l)}}{\partial \mathbf{W}_{ij}^{(l-1)}} , \frac{\partial \boldsymbol{h}^{(l)}}{\partial \mathbf{W}_{ij}^{(l-1)}}  \right\rangle  = \textrm{Var} \left( \mathbf{W}^{(l)} \right) \mathbb{E} \left( \frac{\partial \boldsymbol{s}^{(l-1)}}{\partial \boldsymbol{h}^{(l-1)}} \right)^2 \textrm{Var} \left(  \boldsymbol{s}^{(l-2)} \right)
	\]
	for $i,j \in \mathbb{N}^+$,} where ${\partial \boldsymbol{s}^{(l-1)}} / {\partial \boldsymbol{h}^{(l-1)}}$ adopts the dot operation. Hence, it is easily to prove that ${\partial \boldsymbol{h}^{(l)}}/{\partial \mathbf{W}_{ij}^{(l-1)}} \sim \mathcal{N} (  \boldsymbol{0},   \eta^{\textrm{NTK}}_{l-1} \sigma^2 \mathbf{I}_{n_{l-1}} )$, where $\eta^{\textrm{NTK}}_{l-1} = (n_{l-1} C'_{\phi} C_{\phi})^{-1} $ and $C'_{\phi} = {1}/{\mathbb{E}_{z \sim \mathcal{N}(0,1)} \left( \phi'(z) \right)^2 } $. Moreover, the NTK kernel is defined by 
\[
\begin{aligned}
	K_{\textrm{NTK}}^{(l)} \left(  \boldsymbol{s}^{(l-1)}, \boldsymbol{s}'^{(l-1)} \right) 
	&= K_{\textrm{NTK}}^{(l-1)} \left(  \boldsymbol{s}^{(l-2)}, \boldsymbol{s}'^{(l-2)} \right) \mathbb{E} \left\langle \frac{\partial \boldsymbol{s}^{(l-1)}}{\partial \boldsymbol{h}^{(l-1)}},  \frac{\partial \boldsymbol{s}'^{(l-1)}}{\partial \boldsymbol{h}'^{(l-1)}} \right\rangle \\
	&\quad + K_{\textrm{NNGP}}^{(l)} \left( \boldsymbol{s}^{(l-1)}, \boldsymbol{s}'^{(l-1)} \right)  \quad \text{for $l \geq 2$}
\end{aligned}
\]
with
\[
\lim\limits_{n_{l-1} \to \infty} \mathbb{E} \left\langle \frac{\partial \boldsymbol{h}^{(l)}}{\partial \mathbf{W}_{ij}^{(l-1)}} , \frac{\partial \boldsymbol{h}^{(l)}}{\partial \mathbf{W}_{ij}^{(l-1)}}  \right\rangle  =  \frac{\sigma^2}{C'_{\phi} C_{\phi} }
\]
and
\[
\lim\limits_{n_{l-1} \to \infty} \mathbb{E} \left\langle \frac{\partial \boldsymbol{h}^{(l)}}{\partial \boldsymbol{b}_i^{(l-1)} } , \frac{\partial \boldsymbol{h}^{(l)}}{\partial \boldsymbol{b}_i^{(l-1)} }  \right\rangle  =  \frac{\sigma^2}{C'_{\phi} } \ ,
\]
which is degenerated as $K_{\textrm{NTK}}^{(1)} \left(  \boldsymbol{x}, \boldsymbol{x}' \right) 
= K_{\textrm{NNGP}}^{(1)} \left(  \boldsymbol{x}, \boldsymbol{x}' \right) $ for $l = 1$. It is observed that the NTK are closely related to the gradients and initial parameters.

\subsection{Related Studies}  \label{subsec:rw}
Past decades have witnessed a growing interest in the correspondence between neural network learning and Gaussian processes. Neal et al.~\citep{neal1996:GP} presented the seminal work by showing that a one-hidden-layer network of infinite width turns into a Gaussian process. Cho et al.~\citep{cho2009:GP} linked the multi-layer networks using rectified polynomial activation with compositional Gaussian kernels. Lee et al.~\citep{lee2018:NNGP} showed that the infinitely wide fully connected neural networks with common-used activation functions can converge to Gaussian processes. Recently, the NNGP has been scaled to many types of networks, including Bayesian networks~\citep{novak2018:GP}, deep networks with convolution~\citep{garriga2019:GP}, and recurrent networks~\citep{yang2019:GP}. 

NNGPs can provide a quantitative characterization of how likely certain outcomes are if some aspects of the system are not exactly known. In the experiments of \citep{lee2018:NNGP}, an explicit estimate in the form of variance prediction is given to each test sample. Besides, Pang et al.~\citep{pang2019:NNGP} showed that the NNGP is good at handling data with noise and is superior to discretizing differential operators in solving some linear or nonlinear partial differential equations. Park et al.~\citep{park2020:NNGP} employed the NNGP kernel in the performance measurement of network architectures for the purpose of speeding up the neural architecture search. Pleiss et al.~\citep{pleiss2022:NNGP} leveraged the effects of width on the capacity of neural networks by decoupling the generalization and width of the corresponding NNGP. Despite great progress, numerous studies about NNGP still rely on increasing width to induce the Gaussian processes. Recently, Zhang et al.~\citep{zhang2022:nngp} proposed a depth-induced paradigm that achieves an NNGP by increasing depth, providing complementary support for the existing theory of NNGP.  

The NTK kernel, first proposed by Jacot et al.~\citep{jacot2018:NTK}, relates a neural network trained by randomly initialized gradient descent with a Gaussian distribution. It has been proved that many types of networks, including graph neural networks on bioinformatics datasets~\citep{du2019:GNTK} and convolution neural network~\citep{arora2019:NTK} on medium-scale datasets like UCI database, can derive a corresponding kernel function. Some researchers applied NTK to various fields, such as federated learning~\citep{huang2021:NTK}, mean-field analysis~\citep{mahankali2023:NTK}, and natural language processing~\citep{malladi2023:NTK}. Yang et al. provided considerably comprehensive investigations on NTK~\citep{yang2019:GP}. Recently, Hron et al.~\citep{hron2020:attention} derived the NNGP and NTK from neural networks to multi-head attention architectures as the number of heads tends to infinity. Avidan et al.~\citep{avidan2023:connecting} provided a unified theoretical framework that connects NTK and NNGP using the Markov proximal learning model.

\section{The Unified Kernel} \label{sec:unify}
From Subsection~\ref{subsec:two_kernels}, it is observed that the conventional neural kernels are closely related to the gradients and initial parameters. Inspired by this recognition, this work considers a general form of gradient calculation as follows
{\begin{equation}  \label{eq:lamda}
		\frac{\dif \Theta_t}{\dif t} =  -  \frac{\dif \mathcal{L}(\Theta)}{\dif \Theta} \Big|_t - \lambda \Theta_0 \ ,
	\end{equation}
	where $\mathcal{L}(\Theta)$ denotes the loss function,} $\Theta_0$ indicates the initialized parameter, and $\lambda \in \mathbb{R}$ is the multiplier.  {Eq.~\eqref{eq:lamda} derives an algebraic form for several types of optimization problems, and Subsection~\ref{subsec:corollary} supplements two examples and corresponding corollaries.} Provided Eq.~\eqref{eq:lamda}, we obtain the parameter-updating procedure
\begin{equation} \label{eq:updating_NN}
	\Theta_{t+\dif t} = \Theta_t + \eta \frac{\dif \Theta}{\dif t} = \Theta_t  -  \eta \frac{\dif \mathcal{L}(\Theta)}{\dif \Theta} \Big|_t - \eta \lambda \Theta_0 \ ,
\end{equation}
where $\eta$ indicates the learning rate and $\dif t$ denotes the period unit. Eq.~\eqref{eq:updating_NN}  provides a general perspective for parameter updating; when$\Theta_0$ follows a pre-trained model, the updating parameter takes a trade-off between the original gradient ${\dif \mathcal{L}(\Theta)}/{\dif \Theta}$ and the pre-trained parameter $\Theta_0$ that may maintain well performance. Thus, Eq.~\eqref{eq:updating_NN} conforms to the fine-tuning paradigm. Subsection~\ref{sec:experiments} provides in-depth discussions about Eq.~\eqref{eq:lamda} and Eq.~\eqref{eq:updating_NN}. Besides, it is observed that the multiplier $\lambda$ balances the effectiveness of original gradients ${\dif \mathcal{L}(\Theta)}/{\dif \Theta}$ and initial parameters $\Theta_0$, regardless of the value of $\eta$. Hence, we omit the learning rate in the following theoretical conclusions for greater clarity on the effects of multiplier $\lambda$.  {The algorithmic implementation of Eq.~\eqref{eq:updating_NN} is provided in Subsection~\ref{subsec:experiment_lamda}.}

{\subsection{The Existence of UNK}  \label{subsec:unk}
	In this work, we investigate the limiting characterizations of the inner product of produced variables
	\[
	\left\{~\begin{aligned}
		&\text{Same Timestamps $t$:} \\
		&\quad K_{\textrm{UNK}}^{(l)} \left( t, \boldsymbol{s}^{(l-1)}, \boldsymbol{s}'^{(l-1)} \right)
		\triangleq
		\lim\limits_{n_{l-1} \to \infty} \mathbb{E}_{\Theta_t} \left[   \left\langle \boldsymbol{s}^{(l-1)}(\Theta_{t}) ,\boldsymbol{s}'^{(l-1)} (\Theta_{t}) \right\rangle \right] \ ,\\
		&\text{Different Timestamps $t$ and $0$:} \\
		&\quad K_{\textrm{UNK}}^{(l)} \left( t, 0, \boldsymbol{s}^{(l-1)}, \boldsymbol{s}'^{(l-1)} \right)
		\triangleq
		\lim\limits_{n_{l-1} \to \infty} \mathbb{E}_{\Theta_t,\Theta_0} \left[   \left\langle \boldsymbol{s}^{(l-1)}(\Theta_t) ,\boldsymbol{s}'^{(l-1)} (\Theta_0) \right\rangle \right] \ ,
	\end{aligned}\right.
	\]
	for $l \in [L]$. In the light of the above definition, there is a clear connection to the typical NNGP, if $\Theta_t$ is i.i.d. Gaussian. In this section, we focus on the existence and limiting properties of the $K_{\textrm{UNK}}^{(l)}$ defined above. }  {Now, we present our main conclusion.
	\begin{theorem}  \label{thm:unified}
		For a network of depth $L$ with a Lipschitz activation $\phi$ and in the limit of the layer width $n_1, \dots, n_{L-1} \to \infty$, Eq.~\eqref{eq:lamda} induces a kernel with the following form, for $l\in[L]$ and $t\geq 0$,
		\begin{equation}  \label{eq:our_kernel_1}
			\left\{ \begin{aligned}
				&K_{\textrm{UNK}}^{(l)} \left( t,\boldsymbol{s}^{(l-1)}, \boldsymbol{s}'^{(l-1)} \right)  = \exp\left( \frac{ -\eta ~|\lambda| ~ t}{ \sigma_t^2} \right) \mathbb{E} \left\langle \frac{\partial \boldsymbol{h}^{(l)}}{\partial \Theta_t} , \frac{\partial \boldsymbol{h}'^{(l)}}{\partial \Theta_t}  \right\rangle \ , \\
				&K_{\textrm{UNK}}^{(l)} \left( t, 0, \boldsymbol{s}^{(l-1)}, \boldsymbol{s}'^{(l-1)} \right)  = \exp\left( \frac{ -\eta ~|\lambda| ~ t }{\sqrt{1-\rho_{t}^2} \sigma_0 \sigma_t} \right) \mathbb{E} \left\langle \frac{\partial \boldsymbol{h}^{(l)}}{\partial \Theta_t} , \frac{\partial \boldsymbol{h}'^{(l)}}{\partial \Theta_0}  \right\rangle \ ,
			\end{aligned} \right.
		\end{equation}
		where $\rho_t$ is the correlation multipliers of variables along training epoch $t$, $\sigma_0^2$ and $\sigma_t^2$ denote the variable variances along training epoch 0 and $t$, respectively.  Furthermore, $K_{\textrm{UNK}}(t,\cdot,\cdot)$ has the following properties of limiting kernels
		\begin{itemize}
			\item[(i)] For the case of $\lambda = 0$ or $t=0$, the unified kernel is degenerated as the NTK kernel. Formally, the followings hold for $l \in [L]$
		\end{itemize}
		\[
		\left\{~ \begin{aligned}
			&K_{\textrm{UNK}}^{(l)} \left( t, \boldsymbol{s}^{(l-1)}, \boldsymbol{s}'^{(l-1)} ;\lambda=0 \right) = K_{\textrm{NTK}}^{(l)} \left(  \boldsymbol{s}^{(l-1)}, \boldsymbol{s}'^{(l-1)} \right) \ , \\
			&K_{\textrm{UNK}}^{(l)} \left( t=0, \boldsymbol{s}^{(l-1)}, \boldsymbol{s}'^{(l-1)} \right) = K_{\textrm{NTK}}^{(l)} \left(  \boldsymbol{s}^{(l-1)}, \boldsymbol{s}'^{(l-1)} \right) \ .
		\end{aligned} \right.
		\]
		\begin{itemize}
			\item[(ii)] For the case of $\lambda \neq 0$ and $t \to \infty$, the unified kernel equals to the NNGP kernel, i.e., the following holds for $l\in[L]$ 
		\end{itemize}
		\[
		K_{\textrm{UNK}}^{(l)} \left( t, \boldsymbol{s}^{(l-1)}, \boldsymbol{s}'^{(l-1)} \right) \to K_{\textrm{NNGP}}^{(l)} \left(  \boldsymbol{s}^{(l-1)}, \boldsymbol{s}'^{(l-1)} \right) 
		\quad\text{as}\quad t \to \infty \ .
		\]
\end{theorem} }
Theorem~\ref{thm:unified} presents the existence and  {explicit computations of $K_{\textrm{UNK}}(t,\cdot,\cdot)$ that corresponds to Eq.~\eqref{eq:lamda} for neural kernel learning.} For the case of $t=0$ or $\lambda =0$, the proposed kernel can be degenerated as the NTK kernel, where the parameter updating obeys the Gaussian distribution. For the case of $t \to \infty$ and $\lambda \neq 0$, the proposed kernel can approximate the NNGP kernel well, which implies that a neural network model trained by Eq.~\eqref{eq:lamda} can reach an equilibrium state in a long-time regime. Similar to the NNGP and NTK kernels, the unified kernel is also of a recursive form, that is,
\begin{equation}  \label{eq:our_recursive_1}
	\begin{aligned}
		K_{\textrm{UNK}}^{(l)} \left( t, \boldsymbol{s}^{(l-1)}, \boldsymbol{s}'^{(l-1)} \right)
		&= K_{\textrm{UNK}}^{(l-1)} \left( t,  \boldsymbol{s}^{(l-2)}, \boldsymbol{s}'^{(l-2)} \right) \mathbb{E} \left\langle \frac{\partial \boldsymbol{s}^{(l-1)}}{\partial \boldsymbol{h}^{(l-1)}} ,  \frac{\partial \boldsymbol{s}'^{(l-1)}}{\partial \boldsymbol{h}'^{(l-1)}} \right\rangle \\
		&\quad + \exp\left( \frac{ -\eta ~|\lambda| ~t}{ \sigma_t^2 } \right) K_{\textrm{NNGP}}^{(l)} \left( \boldsymbol{s}^{(l-1)}, \boldsymbol{s}'^{(l-1)}\right) \ .
	\end{aligned}
\end{equation}
The full proof can be accessed in~\ref{app:unified}.

{
	\subsection{Examples and Corollaries}  \label{subsec:corollary}
	This subsection provides two examples and corresponding corollaries for supplementing the computations led by Eqs.~\eqref{eq:lamda} and~\eqref{eq:updating_NN}.}

{
	\paragraph{\bf {Example related to $L_2$ regularizer}} Provided the $L_2$ regularizer, we build the minimization $\min \mathcal{L}(\Theta)  + \gamma (\Theta_0 - \Theta)^2$, which is a standard optimization problem with $L_2$ regularizer. Combined with Eqs.~\eqref{eq:lamda} and~\eqref{eq:updating_NN}, we have the following updating process
	\[
	\Theta_{t+1} = ( 2 \eta \gamma + 1 ) \Theta_t - \eta  \frac{\partial \mathcal{L}(\Theta )}{ \partial \Theta} \Big|_t  - 2 \eta \gamma \Theta_0 \ .
	\]
	Now, we present the following theorem with a UNK kernel for the above equation.
	\begin{corollary}  \label{cor:L_2}
		For a network of depth $L$ with a Lipschitz activation $\phi$ and in the limit of the layer width $n_1, \dots, n_{L-1} \to \infty$, this example induces a kernel with the following form, for $l\in[L]$ and $t\geq 0$,
		\[
		\left\{ \begin{aligned}
			&K_{\textrm{UNK}}^{(l)} \left( t,\boldsymbol{s}^{(l-1)}, \boldsymbol{s}'^{(l-1)} \right)  = C \exp\left( \frac{ -2\eta ~|\lambda| ~ t}{ \sigma_t^2} \right) \mathbb{E} \left\langle \frac{\partial \boldsymbol{h}^{(l)}}{\partial \Theta_t} , \frac{\partial \boldsymbol{h}'^{(l)}}{\partial \Theta_t}  \right\rangle \ , \\
			&K_{\textrm{UNK}}^{(l)} \left( t, 0, \boldsymbol{s}^{(l-1)}, \boldsymbol{s}'^{(l-1)} \right)  = C \exp\left( \frac{ -2\eta ~|\lambda| ~ t }{\sqrt{1-\rho_{t}^2} \sigma_0 \sigma_t} \right) \mathbb{E} \left\langle \frac{\partial \boldsymbol{h}^{(l)}}{\partial \Theta_t} , \frac{\partial \boldsymbol{h}'^{(l)}}{\partial \Theta_0}  \right\rangle \ ,
		\end{aligned} \right.
		\]
		where $C = (2\eta\gamma + 1)^2$.  
	\end{corollary}
	It is evident that the limiting results of Corollary~\ref{cor:L_2} coincide with those of Theorem~\ref{thm:unified}. }

{
	\paragraph{\bf {Example related to $t'$}} From a theoretical perspective, the absolute value of $\lambda$ indicates not only the limiting convergence rate of the UNK kernel but also the optimal search of weight updating. To leverage the effects of various $\lambda$ on the performance of the UNK kernel, we here extend the basic formula to a general case as follows
	\begin{equation}  \label{eq:lamda_t'}
		\frac{\dif \Theta}{\dif t} =  -  \frac{\dif \mathcal{L}(\Theta)}{\dif \Theta} \Big|_t - \lambda \Theta_{t'} \ ,
		\quad\text{for}\quad
		t' \leq t \ .
	\end{equation}
	Obviously, Eq.~\eqref{eq:lamda_t'} has a general updating formulation, taking Eq.~\eqref{eq:lamda} as a special case of $t'=0$. Let $\Theta_{t'}$ indicate a collection of pre-given parameters from pre-training or meta-learning, so that Eq.~\eqref{eq:lamda_t'} becomes an optimization computation for fine-tuning. Hence, the derived kernel may support the theoretical analysis of the fine-tuning learning after pre-training. Provided $\Theta_{t'}$, we can compute the optimal solution $\lambda^*_{t}$ at current epoch stamp $t$ as follows
	\begin{equation}  \label{eq:optimal}
		\lambda^*_t = \arg\min_{\lambda_t} ~ \mathcal{L}(\Theta_{t+\dif t}) - \mathcal{L}(\Theta_t) 
		\quad\text{with}\quad
		\Theta_{t+ \dif t} = \Theta_t  - \eta\frac{\dif \mathcal{L}(\Theta_t)}{\dif \Theta_t} - \eta\lambda_t \Theta_{t-\dif t} \ ,
	\end{equation}
	where $\dif t $ indicates the period unit, set to 1 as default here. This optimization problem can be solved by some mature algorithms, such as Bayesian optimization or grid search. }

{
	Section~\ref{sec:experiments} conducts experiments to verify the effectiveness of Eqs.~\eqref{eq:lamda_t'} and~\eqref{eq:optimal}. Now, we present the following corollary for showing that Eq.~\eqref{eq:lamda_t'} also leads to a UNK kernel.
	\begin{corollary}  \label{cor:unified}
		For a network of depth $L$ with a Lipschitz activation $\phi$ and in the limit of the layer width $n_1, \dots, n_{L-1} \to \infty$, Eq.~\eqref{eq:lamda_t'} induces a kernel with the following form
		\begin{equation}  \label{eq:our_kernel_2}
			K_{\textrm{UNK}}^{(l)} \left( t, t', \boldsymbol{s}^{(l-1)}, \boldsymbol{s}'^{(l-1)} \right) = \exp\left( \frac{ \eta ~|\lambda|~ (t'-t)}{\sqrt{1-\rho_{t,t'}^2}\sigma_{t}\sigma_{t'} } \right) \mathbb{E} \left\langle \frac{\partial \boldsymbol{h}^{(l)}}{\partial \Theta_t} , \frac{\partial \boldsymbol{h}'^{(l)}}{\partial \Theta_{t'}}  \right\rangle \ , \end{equation}
		where $l\in[L]$, $t \geq t'$, $\rho_{t, t'}$ denotes the correlation multiplier of variables along training epochs $t$ and $t'$, and $\sigma_t$ and $\sigma_{t'}$ are the corresponding variances. Furthermore, the unified kernel $K_{\textrm{UNK}}(t,t',\cdot,\cdot)$ has the following properties: 
		\begin{itemize}
			\item[(i)] In the case of $\lambda=0$ or $t=t'$, the unified kernel degenerates as the NTK kernel for $l\in[L]$
			\[
			\begin{aligned}
				& K_{\textrm{UNK}}^{(l)} \left( t, t', \boldsymbol{s}^{(l-1)}, \boldsymbol{s}'^{(l-1)}; \lambda=0 \right) = K_{\textrm{NTK}}^{(l)} \left(  \boldsymbol{s}^{(l-1)}, \boldsymbol{s}'^{(l-1)} \right) \ , \\
				& K_{\textrm{UNK}}^{(l)} \left( t, t, \boldsymbol{s}^{(l-1)}, \boldsymbol{s}'^{(l-1)} \right) = K_{\textrm{NTK}}^{(l)} \left(  \boldsymbol{s}^{(l-1)}, \boldsymbol{s}'^{(l-1)} \right) \ . 
			\end{aligned}
			\]
			\item[(ii)] In the case of $\lambda \neq 0$ and $t-t' \to \infty$, the unified kernel equals to the NNGP kernel for $l\in[L]$
			\[
			K_{\textrm{UNK}}^{(l)} \left( t, t', \boldsymbol{s}^{(l-1)}, \boldsymbol{s}'^{(l-1)} \right) \to K_{\textrm{NNGP}}^{(l)} \left(  \boldsymbol{s}^{(l-1)}, \boldsymbol{s}'^{(l-1)} \right) 
			\quad\text{as}\quad 
			t-t' \to \infty \ .
			\]
		\end{itemize}
	\end{corollary}
	Corollary~\ref{cor:unified}, a general extension of Theorem~\ref{thm:unified}, presents a unified kernel $K_{\textrm{UNK}}(t,t',\cdot,\cdot)$ for neural network learning with Eq.~\eqref{eq:lamda_t'}. For the case of $t=t'$ or $\lambda =0$, the proposed kernel can be degenerated as the NTK kernel, where the parameter updating obeys the Gaussian distribution. Relatively, for the case of $t-t' \to \infty$ and $\lambda \neq 0$, the proposed kernel can approximate the NNGP kernel well, which implies that a neural network model trained by Eq.~\eqref{eq:lamda_t'} can reach an equilibrium state in a long time regime. The key idea of proving Corollary~\ref{cor:unified} is similar to that of Theorem~\ref{thm:unified}. The complete proof of Theorem can be accessed in~\ref{app:t'}. 
}

\section{Convergence and Uniform Tightness}  \label{sec:properties}
We have proposed the UNK kernel, which is built upon the learning dynamics associated with gradient descents and parameter initialization, and have proved that the proposed NUK kernel converges to the NNGP kernel as $t \to \infty$. In this section, we focus on the convergence rate and uniform tightness of the proposed UNK kernel. 

Firstly, the convergence rate corresponds to the smallest eigenvalue of the UNK kernel since the learning convergence is related to the positive definiteness of the limiting neural kernels. It is sufficient to bind the small eigenvalues of $K_{\textrm{UNK}}^{(l)}$ for $l\in[L]$. Here, we consider the neural networks equipped with the ReLU activation and then draw the following conclusion.
\begin{theorem} \label{thm:smallest}
	Let $\boldsymbol{x}_1, \dots, \boldsymbol{x}_N$ be i.i.d. sampled from $P_X$, which satisfies that $P_X = \mathcal{N}(0,\eta^2)$, $\int \boldsymbol{x} \dif P\left(\boldsymbol{x}\right) = 0$, $\int\|\boldsymbol{x}\|_{2} \dif P(\boldsymbol{x}) = \mathbf{\Theta}(\sqrt{n_0})$, and $\int\|\boldsymbol{x}\|_{2}^{2} \dif P(\boldsymbol{x}) = \mathbf{\Theta}(n_0)$. For an integer $r \geq 2$, with probability $1-\delta$, the smallest eigenvalue is bounded, i.e.,
	\[
	\chi_{\min} \left( K_{\textrm{UNK}}^{(l)}  \right) = \mathbf{\Theta}(n_0) 
	\]
	for $l \in [L] $, where $\delta \leq N \exp[ -\Omega(n_0) ] + N^2 \exp[ -\Omega(n_0 N^{-2 /(r-0.5)}) ]$.
\end{theorem}
Theorem~\ref{thm:smallest} provides a tight bound for the smallest eigenvalue of $K_{\textrm{UNK}}^{(l)}$, which is closely related to the training convergence of neural networks. This nontrivial estimation mirrors the characteristics of this kernel, and usually be used as a key assumption for optimization and generalization. Full proof of Theorem~\ref{thm:smallest} is provided in~\ref{app:convergence}.

Secondly, we investigate the uniform tightness of the UNK kernel, which indicates the distribution-dependent convergence of the UNK kernel. Now, we present the following theorem.
\begin{theorem} \label{thm:asymptotic}
	For any $l\in[L]$, the unified kernel $K_{\textrm{UNK}}^{(l)}$ is \textbf{uniformly tight} in $\mathcal{C}(\mathbb{R}^{n_0},\mathbb{R})$.
\end{theorem}
Theorem~\ref{thm:asymptotic} delineates the convergence behavior (i.e., \emph{converges in distribution}) of $K_{\textrm{UNK}}^{(l)}$ for $l\in[L]$, as $t \to \infty$. This reveals an intrinsic characteristic of uniform tightness. Based on Theorem~\ref{thm:asymptotic}, one can obtain the properties of functional limit and continuity of the UNK kernel, in analogy to those of the NNGP kernel~\citep{bracale2020:asymptotic}. To prove the uniform tightness,  it suffices to prove the sequence of $K_{\textrm{UNK}} ( t, \cdot, \cdot)$ led by $t$ is asymptotically tight and $K_{\textrm{UNK}} ( t, \cdot, \cdot)$ is tight for each $t$. The proofs of Theorem~\ref{thm:asymptotic} can be accessed from~\ref{app:tightness}.

\section{Experiments}  \label{sec:experiments}
This section conducts several experiments to evaluate the effectiveness of the proposed UNK kernel. The experiments are performed to discuss the following questions:
\begin{itemize}
	\item[(1)] Whether and how does the multiplier $\lambda$ in Theorem~\ref{thm:unified} affect the convergence of the proposed UNK kernel?
	\item[(2)] What is the representation ability of the UNK kernel? More specifically, what is the difference between the UNK kernels using initial and optimal parameters?
	\item[(3)] How and to what extent does the UNK kernel led by Eq.~\eqref{eq:lamda} improve the performance of pre-trained models?
\end{itemize}

Following the experimental configurations of Lee et al.~\citep{lee2018:NNGP}, we conduct empirical investigations to answer the aforementioned three questions. We focus on a two-hidden-layer MLP trained using Eq.~\eqref{eq:lamda} with various multipliers $\lambda$. The conducted dataset is the \href{http://yann.lecun.com/exdb/mnist/}{MNIST handwritten digit} data, which comprises a training set of 60,000 examples and a testing set of 10,000 examples in 10 classes, where each example is centered in a $28 \times 28$ image. For the classification tasks, the class labels are encoded into an opposite regression formation, where the correct label is marked as 0.9 and the incorrect one is marked as 0.1~\citep{zhang2022:nngp}. Here, we employ 5000 hidden neurons and the \textit{softmax} activation function. Similar to~\citep{arora2019:NNGP}, all weights are initialized with a Gaussian distribution of the mean 0 and variance $0.3/n_l$ for $l \in [L]$. We also force the batch size and the learning rate as 64 and 0.001, respectively. All experiments were conducted on Intel Core-i7-6500U and RTX 4090.

\subsection{Convergence Effects of Various Multipliers $\lambda$}  \label{subsec:experiment_lamda}
From a theoretical perspective, the absolute value of $\lambda$ indicates not only the limiting convergence rate of the UNK kernel but also the optimal search of weight updating. To leverage the effects of various $\lambda$ on the UNK kernel, we here employ the general formula of Eq.~\eqref{eq:lamda_t'} and derive three types of studied models as follows
\begin{equation} \label{eq:training_procedure}
	\left\{\begin{aligned}
		\textrm{Baseline-$\Theta_0$}:&\quad  \frac{\dif \Theta}{\dif t} =  -  \frac{\dif \mathcal{L}(\Theta_t)}{\dif \Theta_t} - \lambda_0 \Theta_0   \ , \\
		\textrm{Baseline-$\Theta_{t'}$}:&\quad  \frac{\dif \Theta}{\dif t} =  -  \frac{\dif \mathcal{L}(\Theta_t)}{\dif \Theta_t} - \lambda_{t'} \Theta_{t'}  \ ,    \\
		\textrm{Grid-$\lambda$}:&\quad  \frac{\dif \Theta}{\dif t} =  -  \frac{\dif \mathcal{L}(\Theta_t)}{\dif \Theta_t} - \lambda^*_t \Theta_{t-\dif t} \ , 
	\end{aligned}\right.
\end{equation}
where \textrm{Baseline-$\Theta_0$} corresponds to initial parameters with preset multiplier $\lambda_0$, \textrm{Baseline-$\Theta_{t'}$} adopts the $t'$-stamp parameters $\Theta_{t'}$ with $t'$-related $\lambda_{t'}$, and \textrm{Grid-$\lambda$} uses the last-stamp parameters $\Theta_{t-1}$ with optimal $\lambda_t^*$ over Eq.~\eqref{eq:optimal}.  {The implementation procedures of Eq.~\eqref{eq:training_procedure} are listed in Algorithm~\ref{alg:gradients}. Here, we employ $\lambda^*_t$ as an indicator for identifying the optimal convergence trajectory of the UNK kernel and set the investigated values of the multiplier $\lambda$ to $\{0.001, 0.01, 0.1, 0, 1, 10\}$. Notice that the optimization problem in Eq.~\eqref{eq:optimal} is solved by grid search with the granularity of 0.001 and 0.01, which are denoted as Grid-0.001 and Grid-0.01, respectively. }
\begin{algorithm}[!htb]
		\caption{The implementation algorithms in Subsection~\ref{subsec:experiment_lamda}}
		\label{alg:gradients}
		\textbf{Input:} Loss function $\mathcal{L}(\Theta)$, learning rate $\eta$, multiplier $\lambda_0$, initial parameters $\Theta_0$, maximum number of iterations $T$ \\
		\textbf{Output:} Optimal parameters $\Theta^*$ \\
		\textbf{Procedures of Baseline-$\Theta_0$:}
		\begin{algorithmic} [1]
			\FOR {$t = 1$ to $T$}
			\STATE Update parameters $\Theta_t \gets \Theta_{t-1} - \eta \cdot ( \nabla_{\Theta} \mathcal{L}(\Theta_{t-1}) + \lambda_0\Theta_0 )$
			\IF{convergence criterion is met}
			\STATE $\Theta^* \gets \Theta_t$; \textbf{break}
			\ENDIF
			\ENDFOR
			\STATE \textbf{return} $\Theta^*$
		\end{algorithmic}
		\textbf{Procedures of Baseline-$\Theta_{t'}$:}
		\begin{algorithmic} [1]
			\STATE Initialize $\lambda_{t'} \gets \lambda_0$ and $\Theta_{t'} \gets \Theta_0$
			\FOR {$t = 1$ to $T$}
			\STATE Update parameters $\Theta_t \gets \Theta_{t-1} - \eta \cdot (  \nabla_{\Theta} \mathcal{L}(\Theta_{t-1}) + \lambda_{t'}\Theta_{t'} )$
			\STATE $s^* \gets \arg\max_{0\leq s \leq t} \{ \mathcal{L}(\Theta_{0}), \dots, \mathcal{L}(\Theta_{t}) \}$
			\STATE Update $\lambda_{t'} \gets \lambda_{s^*}$ and $\Theta_{t'} \gets \Theta_{s^*}$
			\IF{convergence criterion is met}
			\STATE $\Theta^* \gets \Theta_t$; \textbf{break}
			\ENDIF
			\ENDFOR
			\STATE \textbf{return} $\Theta^*$
		\end{algorithmic}
		\textbf{Procedures of Grid-$\lambda$:}
		\begin{algorithmic} [1]
			\STATE Update parameters $\Theta_1 \gets \Theta_0 - \eta \cdot ( \nabla_{\Theta} \mathcal{L}(\Theta_{0}) + \lambda_0\Theta_0 )$
			\FOR {$t = 2$ to $T$}
			\STATE Update parameters $\Theta_t \gets \Theta_{t-1} - \eta \cdot ( \nabla_{\Theta} \mathcal{L}(\Theta_{t-1}) + \lambda_{t-1}\Theta_{t-2} )$
			\IF{convergence criterion is met}
			\STATE $\Theta^* \gets \Theta_t$; \textbf{break}
			\ENDIF
			\FOR{$\lambda \in \{0.001, 0.01, 0.1, 0, 1, 10\}$}
			\STATE $\Theta_{t+ 1}^{\lambda} = \Theta_t  - {\dif \mathcal{L}(\Theta_t)}/{\dif \Theta_t} - \lambda \Theta_{t-1}$
			\ENDFOR
			\STATE Update multiplier $\lambda_t \gets \arg\min_{\lambda} ~ \mathcal{L}(\Theta_{t+1}^{\lambda}) - \mathcal{L}(\Theta_{t}) $
			\ENDFOR
			\STATE \textbf{return} $\Theta^*$
		\end{algorithmic}
\end{algorithm}

\begin{figure*}[t]
	\centering
	\includegraphics[width=1\textwidth]{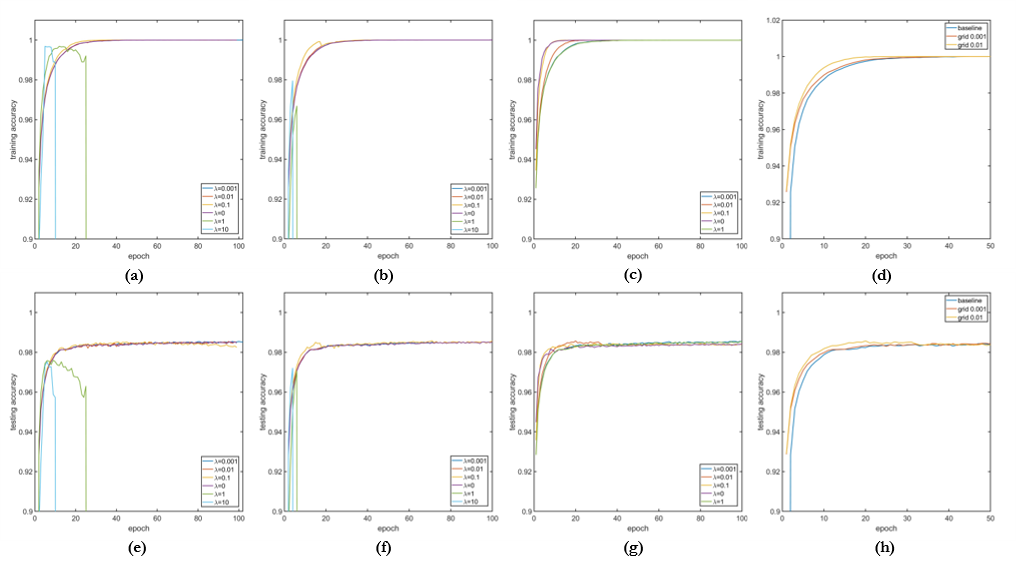}
	\caption{The accuracy curves with various multipliers $\lambda \in \{0.001, 0.01, 0.1, 0, 1, 10\}$, where the x- and y-axes denote the epoch and accuracy, respectively. Training accuracy curves provided (a) Baseline-$\Theta_0$, (b) Baseline-$\Theta_{t'}$, and (c) Grid Search. Testing accuracy curves provided (e) Baseline-$\Theta_0$, (f) Baseline-$\Theta_{t'}$, and (g) Grid Search. Comparison (d) training and (h) testing accuracy curves between Baseline-$\Theta_0$, Grid-0.001, and Grid-0.01. }
	\label{fig:accuracy}
\end{figure*}

Figure~\ref{fig:accuracy} draws various multipliers and the corresponding accuracy curves. There are several observations that (1) the performance of the training procedures in Eq.~\eqref{eq:training_procedure} is comparable to those of typical gradient descent in various configurations as all studied models with apposite $\lambda$ appear convergence, (2) $\lambda=1$ and $\lambda=10$ are too large to hamper the performance of the UNK kernel, and (3) Grid-0.01 provides a starting point for higher accuracy and achieves the fastest convergence speed and best accuracy. The above observations not only show the effectiveness of our proposed UNK kernel, but also coincide with our theoretical conclusions that the UNK kernel converges to the NNGP kernel as $t \to \infty$ and a smaller value of $\lambda$ results in a larger convergence rate.

\begin{table}[t]
	\centering
	\resizebox{1\textwidth}{!}{
		\begin{tabular}{l|l|ll|ll | l|l|ll|ll}
			\toprule
			\textbf{Epoch} & \textbf{Baseline} & \textbf{Grid-0.001} & ~ & \textbf{Grid-0.01} & ~ & \textbf{Epoch} & \textbf{Baseline} & \textbf{Grid-0.001} & ~ & \textbf{Grid-0.01} & ~\\ \midrule
			$\mathbf{t}$ & \textbf{ACC.} & $\mathbf{\lambda^*_t}$ & \textbf{ACC.} & $\mathbf{\lambda^*_t}$ & \textbf{ACC.} & $\mathbf{t}$ & \textbf{ACC.} & $\mathbf{\lambda^*_t}$ & \textbf{ACC.} & $\mathbf{\lambda^*_t}$ & \textbf{ACC.} \\
			\midrule
			1  & 0.1289  & 0.0100  & 0.9257  & 0.0800  & 0.9266  & 14  & 0.9931  & 0.0020  & 0.9943  & 0.0800  & 0.9977  \\ 
			2  & 0.9256  & 0.0020  & 0.9506  & 0.0800  & 0.9521  & 15  & 0.9941  & 0.0020  & 0.9952  & 0.0500  & 0.9984  \\ 
			3  & 0.9504  & 0.0040  & 0.9631  & 0.0900  & 0.9656  & 16  & 0.9949  & 0.0080  & 0.9959  & 0.0700  & 0.9987  \\ 
			4  & 0.9629  & 0.0080  & 0.9708  & 0.0700  & 0.9737  & 17  & 0.9957  & 0.0060  & 0.9966  & 0.0900  & 0.9992  \\ 
			5  & 0.9705  & 0.0070  & 0.9766  & 0.0900  & 0.9793  & 18  & 0.9963  & 0.0070  & 0.9972  & 0.0700  & 0.9995  \\  
			6  & 0.9763  & 0.0050  & 0.9802  & 0.1000  & 0.9839  & 19  & 0.9969  & 0.0070  & 0.9977  & 0.0000  & 0.9996  \\  
			7  & 0.9800  & 0.0060  & 0.9834  & 0.1000  & 0.9870  & 20  & 0.9974  & 0.0100  & 0.9981  & 0.0800  & 0.9998  \\  
			8  & 0.9831  & 0.0000  & 0.9858  & 0.0800  & 0.9899  & 21  & 0.9978  & 0.0070  & 0.9984  & 0.0100  & 0.9997  \\  
			9  & 0.9855  & 0.0080  & 0.9879  & 0.0500  & 0.9922  & 22  & 0.9982  & 0.0100  & 0.9986  & 0.0200  & 0.9999  \\ 
			10  & 0.9875  & 0.0000  & 0.9898  & 0.0900  & 0.9939  & 23  & 0.9984  & 0.0050  & 0.9987  & 0.0000  & 0.9999  \\  
			11  & 0.9896  & 0.0000  & 0.9913  & 0.0600  & 0.9952  & 24  & 0.9986  & 0.0000  & 0.9989  & 0.0000  & 0.9999  \\  
			12  & 0.9910  & 0.0000  & 0.9923  & 0.0600  & 0.9963  & 25  & 0.9988  & 0.0050  & 0.9990  & 0.0000  & 0.9999  \\  
			13  & 0.9922  & 0.0040  & 0.9933  & 0.0700  & 0.9971  & 26  & 0.9989  & 0.0030  & 0.9992  & 0.0000  & 1.0000  \\
			\bottomrule
	\end{tabular} }
	\vspace{0.1cm}
	\caption{Illustration of $\lambda^*_t$ and the corresponding training accuracy (ACC.) of Grid-0.001 and Grid-0.01 over epoch $t$.}
	\label{tab:accuracy_training}
\end{table}
Table~\ref{tab:accuracy_training} further lists the optimal trajectory and the corresponding training accuracy of Grid-0.001 and Grid-0.01 over epoch $t$. It is observed that (1) the optimal trajectory of the UNK kernel and the path of typical gradient descent are not completely consistent, and (2) both Grid-0.001 and Grid-0.01 achieve faster convergence speed and better accuracy than those of the baseline methods. These results further demonstrate the effectiveness of our proposed UNK kernel.

\subsection{Correlation between Initialized and Optimized Parameters}  \label{subsec:fine-tuning}
This experiment investigates the representation ability of our proposed UNK kernel. The indicator is computed as
\[
\gamma^2_i = \frac{ K(T,0,\boldsymbol{x}_i) }{ K(0,0,\boldsymbol{x}_i) K(T,T,\boldsymbol{x}_i) } \ ,
\]
where $\boldsymbol{x}_i$ indicates the $i$-th instance, and $K(T,0,\boldsymbol{x}_i)$ with
\[
K(t,0,\boldsymbol{x}_i) \triangleq K_{\textrm{UNK}}^{(L)} \left( t, \boldsymbol{s}^{L-1}_i(t), {\boldsymbol{s}'_i}^{L-1}(t) \lambda \right) \ .
\]
The value of $\gamma_i$ manifests the \emph{correlation} between outputs of the UNK kernels with initialized and optimized parameters. According to the theoretical results in Section~\ref{sec:unify}, the UNK kernel is said to be \emph{valid} if the kernel outputs brought by initialized and optimized parameters are markedly discriminative. In other words, a valid UNK is able to classify digits well in this experiment, and thus $\gamma_i$ should equal $0.1 \times 1 = 0.1$, where the first 0.1 and 1 denote the accuracy of the UNK with initialized and optimized parameters, respectively. Ideally, the value of $\gamma_i$ in this experiment should trend towards 0.1, that is, $\mathbb{E}_i( \gamma_i ) = 0.1$. If $|\gamma_i| \to 1$, the kernel cannot recognize the difference between outputs brought by initialized and optimized parameters, and thus the kernel is invalid.

\begin{figure*}[t]
	\centering
	\includegraphics[width=1\textwidth]{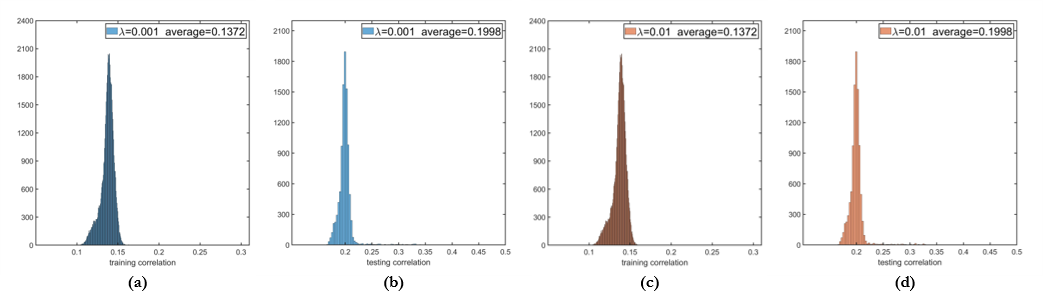}
	\caption{Histograms of training correlation of (a) Grid-0.001 and (c) Grid-0.01, testing correlation of (b) Grid-0.001 and (d) Grid-0.01, where x- and y-axes denote the number of instances and the corresponding correlation, respectively.}
	\label{fig:correlation}
\end{figure*}
Figure~\ref{fig:correlation} displays the (training and testing) correlation histograms and the averages for our proposed UNK kernel with the grid search granularity of 0.001 and 0.01. It is observed that the average training correlation values of Grid-0.001 and Grid-0.01 are almost 0.13 as training accuracy goes to 100\%, which implies that the trained UNK kernel is valid for classifying MNIST. This is a laudable result for the theory and development of neural kernel learning. 

Notice that the average training correlation values for Grid-0.001 and Grid-0.01 are not precisely equal to 0.1, and the average testing correlation values for Grid-0.001 and Grid-0.01 are approximately 0.2 instead of the stated value of 0.1. These discrepancies could be attributed to several factors, including the out-of-distribution errors and the gaps between \textit{softmax} and labeled vectors.

\subsection{Improved Performance on Pre-trained Models}
This experiments focus on the performance of fine-tuning using Eq.~\eqref{eq:lamda} on pre-trained models. The Baseline models contain MLP and LeNet with pre-trained parameters $\Theta_0$, where MLP-$n$ denotes a two-layer MLP with $n$ hidden neurons and MLP-$n$-$m$ indicates a three-layer MLP with $n$ and $m$ hidden neurons. Here, we fine-tune the parameters by computing the gradients of Eq. (3) in each updating step.

Table~\ref{tab:fine_tuning} shows the comparison accuracy pre-trained (Baseline) and fine-tuned (corresponds to the UNK kernel) models with multiplier $\lambda$, the values of which belong to the set $\{0.001, 0.01, 0.1, 0, 1, 10\}$. The best performance is marked in bold, and the symbol $\star$ means that the certain multiplier $\lambda$ is too large to hamper the performance of the UNK kernel. It is observed that fine-tuning using Eq.~\eqref{eq:lamda} with appropriate multipliers indeed improves the accuracy compared to the pre-trained models. This observation further supports the effectiveness of the proposed UNK kernel.

\begin{table}[t]
	\centering
	\caption{Comparison accuracy of pre-trained and fine-tuned models with various multipliers, where the symbol $\star$ means that the certain multiplier $\lambda$ is too large to hamper the performance of the UNK kernel.}
	\label{tab:fine_tuning}
		\begin{tabular}{l|l|l|l|l|l|l|l}
			\toprule
			Models & Baseline & $\lambda=0.001$ & $\lambda=0.01$ & $\lambda=0.1$ & $\lambda=0$ & $\lambda=1$ & $\lambda=10$ \\ \midrule
			MLP-300 &0.8050 & \textbf{0.9726} &\textbf{0.9726} &0.9718 &\textbf{0.9726} & 0.9539& $\star$\\ \midrule
			NN-1000&0.9536 &0.9767 & 0.9706& 0.9706&\textbf{0.9770}&0.9705&$\star$\\ \midrule
			NN-300-100&0.9665& \textbf{0.9750}&0.9742& 0.9702& 0.9744& $\star$ & $\star$\\ \midrule
			LeNet-1& 0.9823& \textbf{0.9884}& 0.9879& 0.9852& 0.9867& $\star$ & $\star$\\ \midrule
			LeNet-5& 0.9805 & 0.9846& \textbf{0.9859}&$\star$ &0.9833&$\star$&$\star$\\ 
			\bottomrule
	\end{tabular} 
\end{table}

\subsection{Discussions}  \label{subsec:discussion}
Here, we consider several more formulas that correspond to Eq.~\eqref{eq:lamda_t'} rather than the basic one in Eq.~\eqref{eq:lamda}. Intuitively, Eq.~\eqref{eq:lamda_t'} takes Eq.~\eqref{eq:lamda} as a special case when one sets $t'=0$. From the theoretical perspective, Eq.~\eqref{eq:lamda_t'} leads to a UNK kernel under mild assumption, which also unifies the limiting properties of the NNGP and NTK kernels. The detailed statement and proof are shown in~\ref{app:t'}. Besides, it is observed that a smaller value of $\lambda$ may lead to a larger convergence rate and the empirical correlation between outputs of the UNK kernels with initialized and optimized parameters approaches the ideal value, which coincides with our results in Theorem~\ref{thm:unified}.

From the empirical perspective, we have shown that Baseline-$\Theta_0$, corresponding to Eq.~\eqref{eq:lamda}, appears convergence and performs comparably to the NTK kernel using a boarder scope of $\lambda$. These experimental results verified the effectiveness of the proposed UNK kernel led by Eq.~\eqref{eq:lamda}. Moreover, Eq.~\eqref{eq:lamda_t'} provides a broader perspective for the implementation of the proposed UNK kernel, from which Baseline-$\Theta_{t'}$ denotes the updating with parameters $\Theta_{t'}$ at a certain time $t'$ and a $t'$-related multiplier $\lambda_{t'}$, and Grid-$\lambda$ employs the optimal $\lambda$ found by grid search. If Baseline-$\Theta_{t'}$ and Grid-$\lambda$ work, it would mean that the UNK kernel learns from a collection of pre-trained parameters $\Theta_{t'}$. From Figure~\ref{fig:accuracy} and Table~\ref{tab:accuracy_training}, both Baseline-$\Theta_{t'}$ and Grid-$\lambda$ perform better than the NTK kernel and Baseline-$\Theta_0$. These empirical results not only further support the effectiveness of the proposed UNK kernel, but also confirm the deep neural kernel can work using pre-trained parameters in contrast to the conventional studies that depend heavily on initial parameters.

This work only investigates the pre-trained parameters obtained by gradient descents. In the future, it is attractive further to explore effective implementations of the proposed UNK kernel and extend the experiment in Subsection~\ref{subsec:fine-tuning} to broader investigations. We conjecture that the pre-trained parameters can be replaced by those obtained from meta learning~\citep{hospedales2021:meta} or transfer learning~\citep{zhuang2020:transfer}. Besides, it is also interesting to extend the theory of deep neural kernels to various fields like neural network learning with knowledge distillation~\citep{gou2021knowledge}.

\section{Conclusions}  \label{sec:conclusions}
In this paper, we proposed the UNK kernel, a unified framework for neural network learning that draws upon the learning dynamics associated with gradient descents and parameter initialization. Our investigation explores theoretical aspects, such as the existence, limiting properties, uniform tightness, and learning convergence of the proposed UNK kernel. Our main findings highlight that the UNK kernel exhibits behaviors akin to the NTK kernel with a finite learning step and converges to the NNGP kernel as the learning step approaches infinity. Experimental results further emphasize the effectiveness of our proposed method.

\section*{Acknowledgment}
This work was supported by the National Science Foundation of China (62406138) and the Natural Science Foundation of Jiangsu Province (BK20230782). 

\newpage
\appendix
\begin{center}
	\Large\textbf{Appendix}
\end{center}
This appendix provides the supplementary materials for our work, constructed according to the corresponding sections therein. 

\section{Full Proof of Theorem~\ref{thm:unified}} \label{app:unified}
{
	All statistics of post-synaptic variables $\boldsymbol{s}$ can be calculated via the moment generating function $\mathcal{M}_{\boldsymbol{s}} (t) = \int \e^{t \boldsymbol{s}}  f(\boldsymbol{s}) \dif \boldsymbol{s}$. Here, we focus on the second moment of $s= \boldsymbol{s}^{(l)}_i$ for $l \in [L]$, that is, 
	\[
	m_2 (s,t) = \int \frac{t^2s^2}{2!}  f(s) \dif s = \int \frac{t^2 s^2(\Theta)}{2!}  ~f_{\Theta}(\Theta) ~\frac{\dif s(\Theta)}{\dif \Theta} \dif \Theta \ .
	\]
	Here, we force $\theta = \Theta^{(l)}$ for $l \in [L]$ and then claim that all element of $\theta$ are i.id. In the above equations, $s$ and $\theta$ denote the variables of hidden states and parameters, respectively. Let $f_{\theta}(\cdot)$ denote the probability density function of any element of $\theta$ and $f_{\theta_t}(\cdot)$ denotes the probability density function of any element of $\theta_t$, where we abbreviate $\theta(t)$ as $\theta_t$ throughout this proof for simplicity. According to the formulation of $m_2 (s)$, we should compute the probability density function $f_{\theta}(\theta)$. Let $\mathrm{Var} ( \theta_{t} ) = \sigma_t^2$, $\dif \theta_t / \dif t$, and $\rho_t$ denotes the correlation multiplier between variables of hidden states $\theta_t$ and $\theta_0$. According to Eq.~\eqref{eq:training_procedure}, we have the general updating formulation
	\[
	\theta_{t+\dif t} = \theta_t + \eta \frac{\dif \theta}{\dif t} = \theta_t  -  \eta \frac{\dif \mathcal{L}(\theta)}{\dif \theta} \Big|_t - \eta \lambda \theta_0 \ ,
	\]
	where $\dif t$ denotes the epoch infinitesimal. Thus, we have 
	\[
	f_{\theta_{t+\dif t}}( u )  = \iiint \delta(  v  ) f_{\theta_t}(x) f_{\nabla_t}(y)  f_{\theta_0} (z) \dif x\! \dif y\! \dif z
	\]
	with
	\[
	\left\{~ \begin{aligned}
		f_{\theta_t}(x) &= \frac{1}{\sigma_t \sqrt{2\pi}} \exp \left( -\frac{x^2}{2\sigma_t^2} \right)   \\
		f_{\nabla_t}(y) &= \frac{1}{\sigma_y \sqrt{2\pi}} \exp \left( -\frac{y^2}{2\sigma_y^2} \right)  \\
		f_{\theta_0} (z) &=  \frac{1}{\sigma_z \sqrt{2\pi}} \exp \left( -\frac{z^2}{2 \sigma_z^2} \right) \\
	\end{aligned}\right.
	\]
	where $v= u- x + y  + \lambda z $, $\nabla_t = {\dif \mathcal{L}(\theta_t)}/{\dif \theta_t}$, and $\delta(\cdot)$ indicates the Dirac-delta function. Moreover, we have
	\[
	\begin{aligned}
		f_{\theta_{t+\dif t}}( u ) &= \iiint \delta(  v  ) f_{\theta_t}(x) f_{\nabla_t}(y)  f_{\theta_0} (z) \dif x\! \dif y\! \dif z \\
		&=\iiint_{v=0} \delta(  v  ) f_{\Theta_t}(x) f_{\nabla_t}(y)  f_{\Theta_0} (z) \dif x\! \dif y\! \dif z \\
		&\quad+ \iiint_{v\neq0} \delta(  v  ) f_{\Theta_t}(x) f_{\nabla_t}(y)  f_{\Theta_0} (z) \dif x\! \dif y\! \dif z \\
		&= \iint_{x,y \in \Omega_{xy}}  f_{\theta_t}(x) f_{\nabla_t}(y) \dif x\! \dif y  \int_{z \in \Omega_z} f_{\theta_0} (z) \dif z \ ,
	\end{aligned}
	\]
	where $\Omega_{xy} = \{ (x,y) \mid (-u+x-y)/ \lambda = 0 \}$, $\Omega_z = \{z \mid z=0 \}$, and the last equality holds from independence.} Thus, we can conjecture that $\theta_{t+\dif t}$ obeys the Gaussian distribution with zero mean. Suppose that
\[
\theta_{t+\dif t} \sim \mathcal{N}(0, \sigma_{t+\dif t}^2)
\quad\text{and}\quad
f_{\theta_{t+\dif t}}(x) = \frac{1}{\sigma_{t+\dif t} \sqrt{2\pi}} \exp \left( -\frac{x^2}{2\sigma_{t+\dif t}^2} \right) \ ,
\]
we have
\[
\begin{aligned}
	m_2 (\theta,t) &= \int \frac{t^2 s^2(\theta)}{2!}  ~f_{\theta}(\theta) ~\frac{\dif s(\theta)}{\dif \theta} \dif \theta \\
	&= \int \frac{t^2 s^2(\theta)}{2!} ~\frac{1}{\sigma_{t+\dif t} \sqrt{2\pi}} \exp \left( -\frac{\theta^2}{2\sigma_{t+\dif t}^2} \right) \frac{\dif s(\theta)}{\dif \theta} \dif \theta \\
	&= \int \frac{t^2 }{2!} \phi^2(h(\theta)) ~\frac{1}{\sigma_{t+\dif t} \sqrt{2\pi}} \exp \left( -\frac{\theta^2}{2\sigma_{t+\dif t}^2} \right) \frac{\dif \phi(h(\theta))}{\dif \theta} \dif \theta  \ ,
\end{aligned}
\]
where $h(\cdot)$ corresponds to $\boldsymbol{h}_i^{(l)}(\cdot)$.  {Inserting $\mathbf{W}^{(l)}$, $\boldsymbol{b}^{(l)}$, and $\boldsymbol{h}^{(l)}$, the above equation can be extended as}
\[
	\begin{aligned}
		& m_2 \left( \mathbf{W}^{(l)},t \right) = \int \frac{t^2 }{2!} \phi^2 \left(\mathbf{W}^{(l)} \boldsymbol{s}^{(l-1)} + \boldsymbol{b}^{(l)} \right) ~\frac{1}{\sqrt{2\pi |\mathbf{\Sigma}_t|} } \exp \left( -\frac{ \mathbf{W}^{(l)}.^2 ~\mathbf{\Sigma}_t^{-1}}{2} \right) \frac{\dif \phi ( \boldsymbol{h}^{(l)} )}{\dif \boldsymbol{h}^{(l)} }  \boldsymbol{s}^{(l-1)}  \dif \mathbf{W}^{(l)} \ , \\
		& m_2 \left( \boldsymbol{b}^{(l)},t \right) = \int \frac{t^2 }{2!} \phi^2 \left(\mathbf{W}^{(l)} \boldsymbol{s}^{(l-1)} + \boldsymbol{b}^{(l)} \right) ~\frac{1}{\sqrt{2\pi |\mathbf{\Sigma}_t|} } \exp \left( -\frac{ \boldsymbol{b}^{(l)}.^2 ~\mathbf{\Sigma}_t^{-1}}{2} \right) \frac{\dif \phi ( \boldsymbol{h}^{(l)} )}{\dif \boldsymbol{h}^{(l)} }  \boldsymbol{1}_{n_l \times 1}  \dif \boldsymbol{b}^{(l)} \ , \\	
		& m_2 \left( \theta,t \right) = \int \frac{t^2 }{2!} \phi^2 \left( \boldsymbol{h}^{(l)} (\theta) \right) ~\frac{1}{\sigma_t \sqrt{2\pi}} \exp \left( -\frac{\theta^2}{2\sigma_t^2} \right) \frac{\dif \phi ( \boldsymbol{h}^{(l)}(\theta) )}{\dif \boldsymbol{h}^{(l)}(\theta) }  \mathbf{W}^{(l)} \frac{\dif \boldsymbol{s}^{(l-1)}(\theta) }{\dif \theta} \dif \theta \ ,
	\end{aligned} 
\]
where $\mathbf{\Sigma}_t$ indicates the corresponding variance matrix. Furthermore, provided two stamps $t$ and $t+\dif t$, we have $\mathbb{E} \left\langle \theta_{t+\dif t}, \theta_t  \right\rangle = m_2( \theta_{t+\dif t}, \theta_t, t+\dif t, t )$ and 
\[
\begin{aligned}
	m_2( \theta_{t+\dif t}, \theta_t, t+\dif t, t ) =&~ \iint \frac{t (t+\dif t)}{2!} \Delta \left( \theta_{t+\dif t}, \theta_t, t+\dif t, t \right) \\
	&\cdot f_{\theta_{t+\dif t},\theta_t} \left( \theta_{t+\dif t}, \theta_t \right) \dif \theta_{t+\dif t} \dif \theta_t \ ,    
\end{aligned}
\]
where
\[
\begin{aligned}
	&\Delta \left( \theta_{t+\dif t}, \theta_t, t+\dif t, t \right) = \\
	&\qquad \left\langle  \phi \left( \boldsymbol{h}^{(l)} (\theta_{t+\dif t}) \right) ,
	\phi \left( \boldsymbol{h}'^{(l)} (\theta_t) \right)   \right\rangle
	\left\langle  \frac{\dif \phi(\boldsymbol{h}^{(l)} \left( \theta_{t+\dif t}) \right)}{\dif \theta_{t+\dif t}} , 
	\frac{\dif \phi(\boldsymbol{h}'^{(l)}  ( \theta_t) }{\dif \theta_t} \right\rangle \ , \\
	& f_{\theta_{t+\dif t},\Theta_t} \left( \theta_{t+\dif t}, \theta_t \right) = \\
	&\qquad
	\frac{1}{2\pi\sqrt{1-\rho_{t+\dif t, t}^2} } \exp \left[ \frac{-1}{2(1-\rho_{t+\dif t, t}^2)} \left( \frac{\theta_{t+\dif t}}{\sigma_{t+\dif t}} - \rho_{t+\dif t,t}\frac{\theta_t}{\sigma_t} \right)^2 \right] \ ,
\end{aligned}
\]
in which $\rho_{t+\dif t, t}$ denotes the correlation multiplier between $\theta_{t+\dif t}$ and $\theta_t$. The estimation of the second moment has been written as a general formula, which can be solved by some mature statistical methods, such as the replica calculation~\citep{mezard1987:Replica}. 

{
	By direct calculations, we can obtain the concerned kernel
	\[
	K_{\textrm{UNK}}^{(l)} \left( t, 0, \boldsymbol{s}^{(l-1)}, \boldsymbol{s}'^{(l-1)} \right)  = \exp\left( \frac{ -\eta~|\lambda| ~t }{\sqrt{1-\rho_t^2}\sigma_{t}\sigma_0} \right) \mathbb{E} \left\langle \frac{\partial \boldsymbol{h}^{(l)}(\theta_t)}{\partial \theta_t} , \frac{\partial \boldsymbol{h}'^{(l)}(\theta_0)}{\partial \theta_0}  \right\rangle \ .
	\]
	and
	\[
	K_{\textrm{UNK}}^{(l)} \left( t, \boldsymbol{s}^{(l-1)}, \boldsymbol{s}'^{(l-1)} \right)  = \exp\left( \frac{ -\eta~|\lambda| ~t }{\sigma_t^2} \right) \mathbb{E} \left\langle \frac{\partial \boldsymbol{h}^{(l)}(\theta_t)}{\partial \theta_t} , \frac{\partial \boldsymbol{h}'^{(l)}(\theta_t)}{\partial \theta_t}  \right\rangle \ .
	\]
	Here, $\boldsymbol{s}^{(l-1)}$ and $\boldsymbol{s}'^{(l-1)}$ are variables led by both $\theta_t$ and $\theta_0$. Similar to the NNGP and NTK kernels, the unified kernel is also of a recursive form as follows
	\[
		\begin{aligned}
			K_{\textrm{UNK}}^{(l)} \left( t, 0, \boldsymbol{s}^{(l-1)}, \boldsymbol{s}'^{(l-1)} \right) =&~ K_{\textrm{UNK}}^{(l-1)} \left( t,  \boldsymbol{s}^{(l-2)}, \boldsymbol{s}'^{(l-2)} \right) \mathbb{E} \left\langle \frac{\partial \boldsymbol{s}^{(l-1)}}{\partial \boldsymbol{h}^{(l-1)}} \Big|_{\theta_t},  \frac{\partial \boldsymbol{s}'^{(l-1)}}{\partial \boldsymbol{h}'^{(l-1)}} \Big|_{\theta_0} \right\rangle \\
			&+ \exp\left( \frac{ -\eta~|\lambda| ~t }{ \sqrt{1-\rho_{t}^2}\sigma_t\sigma_0 } \right) K_{\textrm{NNGP}}^{(l)} \left( \boldsymbol{s}^{(l-1)}(\theta_0), \boldsymbol{s}'^{(l-1)}(\theta_0) \right) 
		\end{aligned} 
	\]
	and
	\[
		\begin{aligned}
			K_{\textrm{UNK}}^{(l)} \left( t, \boldsymbol{s}^{(l-1)}, \boldsymbol{s}'^{(l-1)} \right) =&~ K_{\textrm{UNK}}^{(l-1)} \left( t,  \boldsymbol{s}^{(l-2)}, \boldsymbol{s}'^{(l-2)} \right) \mathbb{E} \left\langle \frac{\partial \boldsymbol{s}^{(l-1)}}{\partial \boldsymbol{h}^{(l-1)}} \Big|_{\theta_t},  \frac{\partial \boldsymbol{s}'^{(l-1)}}{\partial \boldsymbol{h}'^{(l-1)}} \Big|_{\theta_t} \right\rangle \\
			&+ \exp\left( \frac{ -\eta~|\lambda| ~t }{\sigma_t^2} \right) K_{\textrm{NNGP}}^{(l)} \left( \boldsymbol{s}^{(l-1)}(\theta_0), \boldsymbol{s}'^{(l-1)}(\theta_0) \right) \ , \\
		\end{aligned} 
	\] }
The limiting properties of $K_{\textrm{UNK}}^{(l)}$ are similar to those of Corollary~\ref{cor:unified}, which are proved in the next section. $\hfill\square$

\section{Full Proof of Corollary~\ref{cor:unified}}  \label{app:t'}
The key idea of proving Corollary~\ref{cor:unified} is similar to that of Theorem~\ref{thm:unified}. All statistics of post-synaptic variables $\boldsymbol{s}$ can be calculated via the moment generating function $\mathcal{M}_{\boldsymbol{s}} (t) = \int \e^{t \boldsymbol{s}}  f(\boldsymbol{s}) \dif \boldsymbol{s}$. Here, we focus on the second moment of $s= \boldsymbol{s}^{(l)}_i$ for $l \in [L]$ and $i \in [n_l]$, that is, $2! m_2 (s,t) = \int t^2s^2 f(s) \dif s = \int t^2 s^2(\Theta) f_{\Theta}(\Theta) {\dif s(\Theta)}/{\dif \Theta} \dif \Theta$.  {Here, we force $\theta = \Theta^{(l)}$ for $l \in [L]$ and then claim that all element of $\theta$ are i.id. In the above equations, $s$ and $\theta$ denote the variables of hidden states and parameters, respectively. Let $f_{\theta}(\cdot)$ denote the probability density function of any element of $\theta$ and $f_{\theta_t}(\cdot)$ denotes the probability density function of any element of $\theta_t$, where we abbreviate $\theta(t)$ as $\theta_t$ throughout this proof for simplicity. According to the formulation of $m_2 (s)$, we should compute the probability density function $f_{\theta}(\theta)$. Let $\rho_{t,t'}^2$ denote the correlation multiplier between variables of hidden states $\theta_t$ and $\theta_{t'}$. }

{
	According to the introduction in Section~\ref{sec:unify}, Eq.~\eqref{eq:lamda_t'} has a general updating formulation, taking Eq.~\eqref{eq:lamda} as a special case of $t'=0$. Hence, we here take a general formula, i.e., $\theta_{t+ \dif t} = \theta_t  -  \eta{\dif \mathcal{L}(\theta_t)}/{\dif \theta_t}  - \eta\lambda \theta_{t'} $, where $\dif t$ denotes the epoch infinitesimal. Thus, we have $f_{\theta_{t+\dif t}}( u )  = \iiint \delta(  v  ) f_{\Theta_t}(x) f_{\nabla_t}(y)  f_{\theta_0} (z) \dif x\! \dif y\! \dif z $, where $v= u- x + y  + \lambda z $, $\nabla_t = {\dif \mathcal{L}(\theta_t)}/{\dif \theta_t}$, and
	\[
	\left\{~ \begin{aligned}
		f_{\Theta_t}(x) &= \frac{1}{\sigma_t \sqrt{2\pi}} \exp \left( -\frac{x^2}{2\sigma_t^2} \right)  \ , \\
		f_{\nabla_t}(y) &= \frac{1}{\sigma_y \sqrt{2\pi}} \exp \left( -\frac{y^2}{2\sigma_y^2} \right)  \ ,\\
		f_{\theta_0} (z) &=  \frac{1}{\sigma_z \sqrt{2\pi}} \exp \left( -\frac{z^2}{2 \sigma_z^2} \right) \ .\\
	\end{aligned}\right.
	\]
	Moreover, we have $f_{\theta_{t+\dif t}}( u ) = \iint_{x,y\in \Omega_{xy}}  f_{\theta_t}(x) f_{\nabla_t}(y) \dif x\! \dif y  \int_{z \in \Omega_z} f_{\theta_0} (z) \dif z $, where $\Omega_{xy} = \{ (x,y) \mid (-u+x-y)/ \lambda = 0 \}$ and $\Omega_z = \{z \mid z=0 \}$. Thus, we can conjecture that $\theta_{t+\dif t}$ obeys the Gaussian distribution with zero mean.} Suppose that $\theta_{t+\dif t} \sim \mathcal{N}(0, \sigma_{t+\dif t}^2) $ and
\[
f_{\theta_{t+\dif t}}(x) = \frac{1}{\sigma_{t+\dif t} \sqrt{2\pi}} \exp \left( -\frac{x^2}{2\sigma_{t+\dif t}^2} \right) \ .
\]
Thus, we have 
\[
m_2 (\theta,t) = \int \frac{t^2 }{2!} \phi^2(h(\theta)) ~\frac{1}{\sigma_{t+\dif t} \sqrt{2\pi}} \exp \left( -\frac{\theta^2}{2\sigma_{t+\dif t}^2} \right) \frac{\dif \phi(h(\theta))}{\dif \theta} \dif \theta  \ ,
\]
where $h(\cdot)$ corresponds to $\boldsymbol{h}_i^{(l)}(\cdot)$.  {Inserting $\mathbf{W}^{(l)}$, $\boldsymbol{b}^{(l)}$, and $\boldsymbol{h}^{(l)}$, the above equation can be extended as}
\[
	\begin{aligned}
		&m_2 \left( \mathbf{W}^{(l)},t \right) = \int \frac{t^2 }{2!} \phi^2 \left(\mathbf{W}^{(l)} \boldsymbol{s}^{(l-1)} + \boldsymbol{b}^{(l)} \right) ~\frac{1}{\sqrt{2\pi |\mathbf{\Sigma}_t|} } \exp \left( -\frac{ \mathbf{W}^{(l)}.^2 ~\mathbf{\Sigma}_t^{-1}}{2} \right) \frac{\dif \phi ( \boldsymbol{h}^{(l)} )}{\dif \boldsymbol{h}^{(l)} }  \boldsymbol{s}^{(l-1)}  \dif \mathbf{W}^{(l)} \ , \\
		&m_2 \left( \boldsymbol{b}^{(l)},t \right) = \int \frac{t^2 }{2!} \phi^2 \left(\mathbf{W}^{(l)} \boldsymbol{s}^{(l-1)} + \boldsymbol{b}^{(l)} \right) ~\frac{1}{\sqrt{2\pi |\mathbf{\Sigma}_t|} } \exp \left( -\frac{ \boldsymbol{b}^{(l)}.^2 ~\mathbf{\Sigma}_t^{-1}}{2} \right) \frac{\dif \phi ( \boldsymbol{h}^{(l)} )}{\dif \boldsymbol{h}^{(l)} }  \boldsymbol{1}_{n_l \times 1}  \dif \boldsymbol{b}^{(l)} \ , \\
		&m_2 \left( \theta,t \right) = \int \frac{t^2 }{2!} \phi^2 \left( \boldsymbol{h}^{(l)} (\theta) \right) ~\frac{1}{\sigma_t \sqrt{2\pi}} \exp \left( -\frac{\theta^2}{2\sigma_t^2} \right) \frac{\dif \phi ( \boldsymbol{h}^{(l)}(\theta) )}{\dif \boldsymbol{h}^{(l)}(\theta) }  \mathbf{W}^{(l)} \frac{\dif \boldsymbol{s}^{(l-1)}(\theta) }{\dif \theta} \dif \theta \ , 
	\end{aligned}
\]
where $\mathbf{\Sigma}_t$ indicates the corresponding variance matrix. Furthermore, provided two stamps $t$ and $t+\dif t$, we have
\[
{2!} \mathbb{E} \langle \theta_{t+\dif t}, \theta_t  \rangle  = \iint {t (t+\dif t)} \Delta ( \theta_{t+\dif t}, \theta_t, t+\dif t, t ) f_{\theta_{t+\dif t},\theta_t} ( \theta_{t+\dif t}, \theta_t ) \dif \theta_{t+\dif t} \dif \theta_t \ ,
\]
where
\[
\begin{aligned}
	& \Delta \left( \theta_{t+\dif t}, \theta_t, t+\dif t, t \right) = 
	\phi \left( \boldsymbol{h}^{(l)} (\theta_{t+\dif t}) \right)
	\phi \left( \boldsymbol{h}'^{(l)} (\theta_t) \right) 
	\frac{\dif \phi(\boldsymbol{h}^{(l)} \left( \theta_{t+\dif t}) \right)}{\dif \theta_{t+\dif t}} 
	\frac{\dif \phi(\boldsymbol{h}'^{(l)} \left( \theta_t) \right)}{\dif \theta_t}  \ , \\
	& f_{\theta_{t+\dif t},\theta_t} \left( \theta_{t+\dif t}, \theta_t \right) =
	\frac{1}{2\pi\sqrt{1-\rho_{t+\dif t, t}^2} } \exp \left[ \frac{-1}{2(1-\rho_{t+\dif t, t}^2)} \left( \frac{\theta_{t+\dif t}}{\sigma_{t+\dif t}} - \rho_{t+\dif t,t}\frac{\theta_t}{\sigma_t} \right)^2 \right] \ ,
\end{aligned}
\]
in which $\rho_{t+\dif t, t}$ denotes the correlation multiplier between $\theta_{t+\dif t}$ and $\theta_t$. The estimation of the second moment has been written as a general formula, which can be solved by some mature statistical methods, such as the replica calculation~\citep{mezard1987:Replica}. 

By direct calculations, we can obtain the concerned kernel
\[
K_{\textrm{UNK}}^{(l)} \left( t, t', \boldsymbol{s}^{(l-1)}, \boldsymbol{s}'^{(l-1)} \right)  = \exp\left( \frac{ \eta ~|\lambda|~ (t'-t) }{\sqrt{1-\rho_{t,t'}^2}\sigma_{t}\sigma_{t'}} \right) \mathbb{E} \left\langle \frac{\partial \boldsymbol{h}^{(l)}(\theta_t)}{\partial \theta_t} , \frac{\partial \boldsymbol{h}'^{(l)}(\theta_{t'})}{\partial \theta_{t'}}  \right\rangle \ .
\]
Here, $\boldsymbol{s}^{(l-1)}$ and $\boldsymbol{s}'^{(l-1)}$ are variables led by $\theta_t$ and $\theta_{t'}$, respectively. Similar to the NNGP and NTK kernels, the unified kernel is also of a recursive form as follows
\begin{equation}  \label{eq:recursive_app}
		\begin{aligned}
			K_{\textrm{UNK}}^{(l)} \left( t, t', \boldsymbol{s}^{(l-1)}, \boldsymbol{s}'^{(l-1)} \right) =&~ K_{\textrm{UNK}}^{(l-1)} \left( t, t',  \boldsymbol{s}^{(l-2)}, \boldsymbol{s}'^{(l-2)} \right) \mathbb{E} \left\langle \frac{\partial \boldsymbol{s}^{(l-1)}}{\partial \boldsymbol{h}^{(l-1)}} \Big|_{\theta_t},  \frac{\partial \boldsymbol{s}'^{(l-1)}}{\partial \boldsymbol{h}'^{(l-1)}} \Big|_{\theta_{t'}} \right\rangle \\
			&+ \exp\left( \frac{ \eta~|\lambda|~ (t'-t)}{ \sqrt{1-\rho_{t,t'}^2}\sigma_{t}\sigma_{t'} } \right) K_{\textrm{NNGP}}^{(l)} \left( \boldsymbol{s}^{(l-1)}(\theta_0), \boldsymbol{s}'^{(l-1)}(\theta_0) \right) \ .
		\end{aligned}
\end{equation}

{
	Next, we will analyze the limiting properties of $K_{\textrm{UNK}}^{(l)}$. It is observed that
	\[
	\left\{ \begin{aligned}
		\exp\left( \frac{ \eta~|\lambda=0|~ (t'-t) }{\sqrt{1-\rho_{t,t'}^2}\sigma_{t}\sigma_{t'}} \right) = 1 \ , &\quad\text{in the case of $\lambda=0$} \ , \\
		\exp\left( \frac{\eta ~|\lambda|~ (t-t') }{\sqrt{1-\rho_{t,t'}^2}\sigma_{t}\sigma_{t'}} \right) = 1 \ , &\quad\text{in the case of $\lambda \neq 0$ and $t=t'$}  \ .
	\end{aligned} \right.
	\]
	Therefore, Eq.~\eqref{eq:our_kernel_1} is degenerated as the NTK kernel in the case of (i-a) $\lambda=0$ or (i-b) $\lambda \neq 0$ and $t=t'$. } In the case of $\lambda \neq 0$, we have
\[
\lim\limits_{t-t' \to \infty} K_{\textrm{UNK}}^{(l)} \left( t, t', \boldsymbol{s}^{(l-1)}, \boldsymbol{s}'^{(l-1)} \right) \to K_{\textrm{NNGP}}^{(l)} \left( \boldsymbol{s}^{(l-1)}, \boldsymbol{s}'^{(l-1)} \right) 
\quad\text{as}\quad
t-t' \to \infty \ .
\]
According to Eq.~\eqref{eq:recursive_app}, one has
\[
	\begin{aligned}
		\int_{t'}^t & K_{\textrm{UNK}}^{(l)}  \left( t, t', \boldsymbol{s}^{(l-1)}, \boldsymbol{s}'^{(l-1)} \right) \dif t \\
		=&~ \int_{t'}^t K_{\textrm{UNK}}^{(l-1)} \left( t, t',  \boldsymbol{s}^{(l-2)}, \boldsymbol{s}'^{(l-2)} \right) \mathbb{E} \left\langle \frac{\partial \boldsymbol{s}^{(l-1)}}{\partial \boldsymbol{h}^{(l-1)}} \Big|_{\theta_t},  \frac{\partial \boldsymbol{s}'^{(l-1)}}{\partial \boldsymbol{h}'^{(l-1)}} \Big|_{\theta_{t'}} \right\rangle \dif t \\
		&+ \int_{t'}^t \exp\left( \frac{ (t'-t) ~|\lambda|}{\sqrt{1-\rho_{t,t'}^2}\sigma_{t}\sigma_{t'}} \right) K_{\textrm{NNGP}}^{(l)} \left( \boldsymbol{s}^{(l-1)}(\theta_t), \boldsymbol{s}'^{(l-1)}(\theta_{t'}) \right) \dif t \\
		=&~ \frac{\sqrt{1-\rho_{t,t'}^2}\sigma_{t}\sigma_{t'}}{|\lambda|} \int_{t'}^t \left[  \exp\left( \frac{ (t'-t) ~|\lambda|}{\sqrt{1-\rho_{t,t'}^2}\sigma_{t}\sigma_{t'}} \right) K_{\textrm{NNGP}}^{(l)} \left( \boldsymbol{s}^{(l-1)}(\theta_t), \boldsymbol{s}'^{(l-1)}(\theta_{t'}) \right)  \right]_{\partial t}  \dif t \ ,
	\end{aligned} 
\]
where $[\cdot]_{\partial t}$ denotes the differential operation with respect to $t$. Thus, for any $t'$, it is easy to prove that 
\[
\lim\limits_{t\to\infty} \int_{t'}^t K_{\textrm{UNK}}^{(l)} \left( t, t', \boldsymbol{s}^{(l-1)}, \boldsymbol{s}'^{(l-1)} \right) \dif t = \frac{\sqrt{1-\rho_{t,t'}^2}\sigma^2}{\eta|\lambda|} K_{\textrm{NNGP}}^{(l)} \left( \boldsymbol{s}^{(l-1)}, \boldsymbol{s}'^{(l-1)} \right) \ .
\]
Here, we consider that the correlation multiplier $\rho_{t,t'}$ is negatively proportional to $t-t'$ since the variable correlation becomes smaller as the stamp gap increases. Generally, we employ $\rho_{t,t'} = \mathbf{\Theta} ( (t-t')^{-1} )$ and $\lim\limits_{t-t' \to \infty} {\rho_{t,t'}}/(t-t') = C \in \mathbb{R}$. Thus, one has 
\[
\lim\limits_{t-t' \to \infty} K_{\textrm{UNK}}^{(l)} ( t, t', \boldsymbol{s}^{(l-1)}, \boldsymbol{s}'^{(l-1)} ) = K_{\textrm{NNGP}}^{(l)} (  \boldsymbol{s}^{(l-1)}, \boldsymbol{s}'^{(l-1)} ) \ ,
\]
in which we omit the constant multiplier. From $\sqrt{1-\rho_{t,t'}^2}\sigma_{t}\sigma_{t'} \to \max_t \sigma_t^2$ as $t-t' \to \infty$, we can complete the proof. $\hfill\square$

{
	\paragraph{\textbf{For the case of $\lambda=0$}} 
	For the case of $\lambda=0$, we can update $\theta$ from $\theta_{t+ \dif t} = \theta_t  -  \eta{\dif \mathcal{L}(\theta_t)}/{\dif \theta_t}$. Let $f_{\theta_t}(\cdot)$ denote the probability density function of $\theta(t)$. Thus, we have $f_{\theta_{t+\dif t}}( u )  = \iint \delta(  v  ) f_{\theta_t}(x) f_{\nabla_t}(y)  \dif x\! \dif y$, where $v= u- x + y$, $\nabla_t = {\dif \mathcal{L}(\theta_t)}/{\dif \theta_t}$, 
	\[
	f_{\theta_t}(x) = \frac{1}{\sigma_x \sqrt{2\pi}} \exp \left( -\frac{x^2}{2\sigma_x^2} \right) \ ,
	\quad\text{and}\quad
	f_{\nabla_t}(y) = \frac{1}{\sigma_y \sqrt{2\pi}} \exp \left( -\frac{y^2}{2\sigma_y^2} \right)  \ .
	\]
	Thus, it is feasible to conjecture that $\theta_{t+\dif t}$ obeys the Gaussian distribution with zero mean. We define $\theta_{t+\dif t} \sim \mathcal{N}(0, \sigma_u^2) $. Hence, the second moment in $\mathcal{M}_{\boldsymbol{s}} (\cdot)$ becomes 
	\[
	m_2 (s) = \int s^2  ~f(s) \dif s = \int s^2(\theta) ~\frac{1}{\sigma_u \sqrt{2\pi}} \exp \left( -\frac{s^2(\theta)}{2\sigma_u^2} \right) \frac{\dif s(\theta)}{\dif \theta} \dif \theta \ ,
	\]
	where $s= \boldsymbol{s}^{(l)}_i$ for $l \in [L]$ and $i \in [n_l]$.} Based on the above equations, we can obtain the concerned kernel
\[
K_{\textrm{UNK}}^{(l)} \left( t, t', \boldsymbol{s}^{(l-1)}, \boldsymbol{s}'^{(l-1)} ; \lambda=0 \right) = \mathbb{E} \left\langle \frac{\partial \boldsymbol{h}^{(l)}(\theta_t)}{\partial \theta_t} , \frac{\partial \boldsymbol{h}'^{(l)}(\theta_{t'})}{\partial \theta_{t'}}  \right\rangle  \ ,
\]
which coincides with the theory of NTK and our proposed unified kernel. $\hfill\square$

\section{Tight Bound for Convergence} \label{app:convergence}
The key idea of proving Theorem~\ref{thm:smallest} is based on the following inequalities about the smallest eigenvalue of real-valued symmetric square matrices. Given two symmetric matrices $\mathbf{A}, \mathbf{B}\in\mathbb{R}^{m \times m}$, it is observed that 
\[
\chi_{\min}(\mathbf{A}\mathbf{B}) \geq \chi_{\min}(\mathbf{A}) \cdot \min_{i\in[m]} \mathbf{B}(i,i)
\quad\text{and}\quad
\chi_{\min}(\mathbf{A}+\mathbf{B}) \geq \chi_{\min}(\mathbf{A}) + \chi_{\min}(\mathbf{B}) \ .
\]
From Eq.~\eqref{eq:our_recursive_1}, we can unfold the UNK kernel as a sum of covariance of the sequence of random variables $\{\boldsymbol{s}^{(l-1)}\}$. Thus, we can bound $\chi_{\min} ( K_{\textrm{UNK}}^{(l)} )$ by $\mathrm{Cov}(\boldsymbol{s}^{(l-1)},\boldsymbol{s}^{(l-1)})$ via a chain of feed-forward compositions in Eq.~\eqref{eq:feedforward}.

We begin this proof with the following lemmas.
\begin{lemma} \label{lemma:concentration}
	Let $f:\mathbb{R}^{n_0} \to \mathbb{R}$ be a Lipschitz continuous function with constant $C_{n_0}$ and $P_X$ denote the Gaussian distribution $\mathcal{N}(0,\eta^2)$, then for $\forall~ \delta>0$, there exists $c >0$, s.t.
	\begin{equation} \label{eq:log}
		\mathbb{P} \left( \left| f(\boldsymbol{x})-\int f \left(\boldsymbol{x}^{\prime}\right) \dif P_{X}\left(\boldsymbol{x}^{\prime}\right) \right| > \delta \right) \leq 2 \exp \left( \frac{-c \delta^{2}}{C_{n_0}^2} \right) \ .
	\end{equation}
\end{lemma}
Lemma~\ref{lemma:concentration} shows that the Gaussian distribution corresponding to our samples satisfies the log-Sobolev inequality~\eqref{eq:log}, with some constants unrelated to dimension $n_0$. This result also holds for the uniform distributions on the sphere or unit hypercube~\citep{nguyen2021:eigenvalues}. 

\begin{lemma} \label{lemma:input}
	Suppose that $\boldsymbol{x}_1, \dots, \boldsymbol{x}_N$ are i.i.d. sampled from $\mathcal{N}(0,\eta^2)$, then with probability $1-\delta>0$, we have $\|\boldsymbol{x}_i\|_2 = \mathbf{\Theta}(\sqrt{n_0})$ and $ |\langle \boldsymbol{x}_i,\boldsymbol{x}_j \rangle|^r \leq n_0 N^{-1/(r-0.5)} $ for $i \neq j$, where $\delta \leq N \exp\left[ -\Omega(n_0) \right] + N^{2} \exp \left[ -\Omega\left(n_0 N^{-2 /(r-0.5)}\right) \right]$.
\end{lemma}
From Definition 1 of the manuscript, we have $\int\|\boldsymbol{x}\|_{2}^{2} \dif P_X(\boldsymbol{x}) = \mathbf{\Theta}(n_0)$. Since $\boldsymbol{x}_1, \dots, \boldsymbol{x}_n$ are i.i.d. sampled from $P_X = \mathcal{N}(0,\eta^2)$, for $\forall$ $i\in[N]$, we have $\|\boldsymbol{x}_i\|_2^2 = \mathbf{\Theta}(n_0)$ with probability at least $1- N\exp[\Omega(n_0)]$. Provided $\boldsymbol{x}_i$, the single-sided inner product $\langle \boldsymbol{x}_i,\cdot \rangle$ is Lipschitz continuous with the constant $C_{n_0} = \mathcal{O}(\sqrt{n_0})$. From Lemma~\ref{lemma:concentration}, for $\forall~ j\neq i$, we have $\mathbb{P} \left( |\langle \boldsymbol{x}_i,\boldsymbol{x}_j \rangle| > \delta \right) \leq 2 \exp( -\Omega\left( {\delta}^2\right)/C_{n_0}^2 ) $. Then, for $r \geq 2$, we have $ \mathbb{P} ( \max_{j\neq i} |\langle \boldsymbol{x}_i,\boldsymbol{x}_j \rangle|^r > \delta ) \leq N^{2} \exp[ -\Omega( {\delta}^2 )] $. We complete the proof by setting $\delta \leq n_0N^{-1/(r-0.5)}$. 
$\hfill\square$

\subsection{Full proof of Theorem~\ref{thm:smallest}}
We start this proof with some notations. For convenience, we force $n =|\boldsymbol{s}^{(1)}|_{\#} = |\boldsymbol{s}^{(2)}|_{\#} = \dots = |\boldsymbol{s}^{(L)}|_{\#}$ or equally $n=n_1=\dots=n_L$. We also abbreviate the covariance $\mathrm{Cov}(\boldsymbol{s}^{(l)},\boldsymbol{s}^{(l)})$ as $\mathbf{C}_{l}$ throughout this proof. By unfolding the $K_{\textrm{UNK}}^{(l)}$ kernel equation that omits the epoch stamp, i.e., $K_{\textrm{UNK}}^{(l)} (\boldsymbol{x}_i,\boldsymbol{x}_j) = \mathbb{E}[\langle f(\boldsymbol{x}_i; \boldsymbol{\theta}), f(\boldsymbol{x}_j; \boldsymbol{\theta})\rangle]$ for $\boldsymbol{x}_i, \boldsymbol{x}_j \in \mathcal{D} $, we have
\begin{equation} \label{eq:K_cov}
	K_{\textrm{UNK}}^{(l)} (\boldsymbol{x}_i, \boldsymbol{x}_j) = \frac{1}{M_{\boldsymbol{z}}} \left[ \sum_{\kappa} \varphi_{\kappa} + \sum_{\kappa_1 \neq \kappa_2} \phi_{\kappa_1,\kappa_2} \right] \ ,
\end{equation}
where $\varphi_l = \mathbb{E} [ \langle \boldsymbol{s}^{l}, \boldsymbol{s}^{(l)} \rangle ]$ and $\psi_{l_1l_2} = \sum\nolimits_{p,q} \mathbb{E} [ \boldsymbol{s}_p^{(l_1)} \boldsymbol{s}_q^{(l_2)} ]$ for $l_1 \neq l_2$, in which the subscript $p$ indicates the $p$-th element of vector $\boldsymbol{s}^{(l)}$. From Theorem 1 of the manuscript, the sequence of random variables $\boldsymbol{s}^{(l)}$ is weakly dependent with $\beta(t) \to \infty$ as $t\to \infty$. Thus, $\psi_{l_1l_2}$ is an infinitesimal with respect to $|l_2-l_1|$ when $l_1 \neq l_2$. Invoking $\chi_{\min}(\mathbf{P}\mathbf{Q}) \geq \chi_{\min}(\mathbf{P}) \min_{i\in[m]} \mathbf{Q}(i,i) $ and $\chi_{\min}(\mathbf{P}+\mathbf{Q}) \geq \chi_{\min}(\mathbf{P}) + \chi_{\min}(\mathbf{Q}) $ into Eq.~\eqref{eq:K_cov}, we have $\chi_{\min}( K_{\textrm{UNK}}^{(l)} ) \geq \sum\nolimits_{l} \chi_{\min} ( \mathbf{C}_{l} )$ and $\chi_{\min} ( \mathbf{C}_{l} ) \geq \chi_{\min} ( \mathbf{C}_{l} )$ for $l \in [L]$. By iterating the above inequalities, we have $\chi_{\min} ( K_{\textrm{UNK}}^{(l)} ) \geq \sum_l \chi_{\min}( \mathbf{C}_1 ) $. From the Hermite expansion~\citep{zhang2021:arise} of the ReLU function, we have $	\mu_{r}(\psi) = (-1)^{\frac{r-2}{2}} (r-3)!! / \sqrt{2 \pi r!}$, where $r \geq 2$ indicates the expansion order. Thus, we have
\begin{equation} \label{eq:thm2_4}
	\begin{aligned}
		\chi_{\min}\left( \mathbf{C}_1 \right) &= \chi_{\min}\left( \psi(\mathbf{W}^{(1)}\mathbf{X}) \psi(\mathbf{W}^{(1)}\mathbf{X})^{\top} \right)
		\geq \mu_{r}(\phi)^2 \chi_{\min}\left( \mathbf{X}^{(r)} \left( \mathbf{X}^{(r)} \right)^{\top} \right) \\
		&\geq \mu_{r}(\psi)^2 \left( \min_{i\in[N]} \|\boldsymbol{x}_i\|_2^{2r} - (N-1) \max_{j\neq i} |\langle \boldsymbol{x}_i,\boldsymbol{x}_j \rangle|^r \right) 
		\geq \mu_{r}(\psi)^2 ~\Omega(n_0) \ ,
	\end{aligned}
\end{equation}
where $\mathbf{X}=[\boldsymbol{x}_1, \dots,\boldsymbol{x}_N]$, the superscript $(r)$ denotes the $r$-th Khatri Rao power of the matrix $\mathbf{X}$, the first inequality follows from $	\mu_{r}(\psi) $, the second one holds from Gershgorin Circle Theorem~\citep{salas1999gershgorin}, and the third one follows from Lemma~\ref{lemma:input}. Therefore, we can obtain the lower bound of the smallest eigenvalue by plugging Eq.~\eqref{eq:thm2_4} into $\chi_{\min} ( K_{\textrm{UNK}}^{(l)} )$. On the other hand, it is observed from Lemma~\ref{lemma:tightness} that for $l \in [L]$,
\begin{equation} \label{eq:thm2_5}
	\left\{~\begin{aligned}
		& \| \boldsymbol{s}_p^{(l)} \|^2_2 = \mathbb{E}_{\mathbf{W}^{(l)}_p} \left[ \psi(\mathbf{W}^{(l)}_p\boldsymbol{s}^{(l-1)})^2 \right] = \| \boldsymbol{s}_q^{(l)} \|^2, \quad\text{for}\quad \forall q \neq p ,\\
		&\| \boldsymbol{s}^{(l)} \|_2^2 = \mathbb{E}_{\mathbf{W}^{(l)}} \left[ \psi(\mathbf{W}^{(l)}\boldsymbol{s}^{(l-1)})^2 \right] \leq \| \boldsymbol{s}^{(l)} \|_2^2 \ . 
	\end{aligned}\right.
\end{equation}
Thus, we have $\chi_{\min}( K_{\textrm{UNK}}^{(l)} ) \leq N^{-1}{\tr( K_{\textrm{UNK}}^{(l)} )} = N^{-1} \sum_{i}^{N} K_{\textrm{UNK}}^{(l)}(\boldsymbol{x}_i, \boldsymbol{x}_i) \leq \mathbf{\Theta}(n_0) $, where the last inequality holds from Lemma~\ref{lemma:input}. This completes the proof. $\hfill\square$

\section{Uniform Tightness of $K_{\textrm{UNK}}^{(l)}$} \label{app:tightness}
To prove the uniform tightness,  it suffices to prove the sequence of $K_{\textrm{UNK}} ( t, \cdot, \cdot)$ led by $t$ is asymptotically tight and $K_{\textrm{UNK}} ( t, \cdot, \cdot)$ is tight for each $t$. Hence, Theorem~\ref{thm:asymptotic} establishes upon three useful lemmas from~\citep{zhang2022:nngp}.
\begin{lemma} \label{lemma:tightness}
	Let $\{\boldsymbol{s}_1, \boldsymbol{s}_2, \dots, \boldsymbol{s}_t\}$ denote a sequence of random variables in $\mathcal{C}(\mathbb{R}^{n_0},\mathbb{R})$. This stochastic process is \textbf{uniformly tight} in $\mathcal{C}(\mathbb{R}^{n_0},\mathbb{R})$, if and only if the following two facts hold:
	(1) $\boldsymbol{x}=\boldsymbol{0}$ is a uniformly tight point of $\boldsymbol{s}_t(\boldsymbol{x})$ ($t \in [T]$) in $\mathcal{C}(\mathbb{R}^{n_0},\mathbb{R})$; 
	(2) for any $\boldsymbol{x}, \boldsymbol{x}' \in \mathbb{R}^{n_0}$, and $t \in [T]$, there exist $\alpha, \beta, C >0$, such that $\mathbb{E} \left[ | \boldsymbol{s}_t(\boldsymbol{x}) - \boldsymbol{s}_t(\boldsymbol{x}') |^{\alpha} \right] \leq C \| \boldsymbol{x} - \boldsymbol{x}' \|_{\beta+n_0} $.
\end{lemma}
Lemma~\ref{lemma:tightness} shows core guidance for proving Theorem~\ref{thm:asymptotic}, in which $K_{\textrm{UNK}} ( t, \cdot, \cdot)$ is uniformly tight if and only if it is asymptotically tight and $K_{\textrm{UNK}} ( t, \cdot, \cdot)$ is tight for each $t$. Lemma~\ref{lemma:tightness_1} and Lemma~\ref{lemma:tightness_2} prove these two facts in Lemma~\ref{lemma:tightness}, respectively.

\begin{lemma} \label{lemma:tightness_1}
	Based on the notations of Lemma~\ref{lemma:tightness}, $\boldsymbol{x}=\boldsymbol{0}$ is a uniformly tight point of $\boldsymbol{s}_t(\boldsymbol{x})$ ($t \in [T]$) in $\mathcal{C}(\mathbb{R}^{n_0},\mathbb{R})$.
\end{lemma}
\begin{lemma} \label{lemma:tightness_2}
	Based on the notations of Lemma~\ref{lemma:tightness}, for any $\boldsymbol{x}, \boldsymbol{x}' \in \mathbb{R}^{n_0}$ and $t \in [T]$, there exist $\alpha, \beta, C >0$, such that $\mathbb{E} \left[ \| \boldsymbol{s}_t(\boldsymbol{x}) - \boldsymbol{s}_t(\boldsymbol{x}') \|_{\alpha}^{\textrm{sup}} ~\right] \leq C \| \boldsymbol{x} - \boldsymbol{x}' \|_{\beta+n_0}$.
\end{lemma}
The proofs of lemmas above can be accessed from~\ref{app:tightness}. Notice that the above lemmas take the stochastic process of hidden neuron vectors with increasing epochs regardless of the layer index, i.e., the above lemmas hold for $\boldsymbol{s}^{(l)}$ $(l\in[L])$. For the case of two stamps $t$ and $t'$ where $t' \ll t$, the concerned stochastic process becomes $\{\boldsymbol{s}_{t'}, \boldsymbol{s}_2, \dots, \boldsymbol{s}_t\}$, and thus the above conclusions also hold. Therefore, Theorem~\ref{thm:asymptotic} can be completely proved by invoking Lemmas~\ref{lemma:tightness_1} and~\ref{lemma:tightness_2} into Lemma~\ref{lemma:tightness}.

\subsection{Full Proof of Lemma~\ref{lemma:tightness}}
Lemma~\ref{lemma:tightness} can be straightforwardly derived from Kolmogorov Continuity Theorem~\citep{stroock1997:Kolmogorov}, provided the Polish space $(\mathbb{R}, |\cdot|)$. 

\subsection{Full Proof of Lemma~\ref{lemma:tightness_1}}
It suffices to prove that (1) $\boldsymbol{x}=\boldsymbol{0}$ is a tight point of $\boldsymbol{s}_t(\boldsymbol{x})$ ($t \in [T]$) in $\mathcal{C}(\mathbb{R}^{n_0},\mathbb{R})$. This conjecture is self-evident since every probability measure in $(\mathbb{R}, |\cdot|)$ is tight~\citep{zhang2021:arise}. (2) The statistic $(\boldsymbol{s}_1(\boldsymbol{0})+ \dots + \boldsymbol{s}_t(\boldsymbol{0})) / t$ converges in distribution as $t \to \infty$. This conjecture has been proved by Corollary~\ref{cor:unified}. We finish the proof of this lemma.  $\hfill\square$

\subsection{Full Proof of Lemma~\ref{lemma:tightness_2}}
This proof follows mathematical induction. Before that, we show the following preliminary result. Let $\theta$ be one element of the augmented matrix $(\mathbf{W}^{(l)}, \boldsymbol{b}^{(l)})$ at the $l$-th layer, then we can formulate its characteristic function as $\varphi(t) = \mathbb{E}\left[ \e^{\mathrm{i}\theta t} \right] = \e^{-\eta^2 t^2/2}$ with $\theta \sim \mathcal{N}(0,\eta^2)$, where $\mathrm{i}$ denotes the imaginary unit with $\mathrm{i} = \sqrt{-1}$. Thus, the variance of hidden random variables at the $l$-th layer becomes
\begin{equation} \label{eq:sigma}
	\sigma^2_l = \eta^2 \left[ 1 + \frac{1}{n_l}  \big\| \varphi \circ \boldsymbol{s}^{(l-1)} \big\| \right] \ .
\end{equation}

Next, we provide two useful definitions from~\citep{zhang2022:nngp}.
\begin{definition} \label{def:well_posed}
	A  function $\phi:\mathbb{R}\to\mathbb{R}$ is said to be \textbf{well-posed}, if $\phi$ is first-order differentiable, and its derivative is bounded by a certain constant $C_{\phi}$. In particular, the commonly used activation functions like ReLU, \textit{tanh}, and \textit{sigmoid} are well-posed (see Table~\ref{tab:activation}). Further, a matrix $\mathbf{W}$ is said to be \textbf{stable-pertinent} for a well-posed activation function $\phi$, in short $\mathbf{W} \in SP(\phi)$, if the inequality $C_{\phi} \|\mathbf{W}\| < 1$ holds.
\end{definition}
\begin{table}[!htb]
	\centering
	\caption{Well-posedness of the commonly-used activation functions.}
	\label{tab:activation}
	\begin{tabular}{l|l}
		\toprule
		Activations $\phi$ & Well-Posedness  \\ \midrule
		ReLU & $\|\phi'(\boldsymbol{x})\| \leq 1$ \\ \midrule
		$\tanh$     & $\|\phi'(\boldsymbol{x})\| = \| 1- \sigma^2(\boldsymbol{x}) \| \leq 1$  \\ \midrule
		sigmoid  & $\|\phi'(\boldsymbol{x})\| = \| \phi(\boldsymbol{x})(1- \phi(\boldsymbol{x})) \| \leq 0.25$  \\ \bottomrule
	\end{tabular} 
\end{table}

Since the activation $\phi$ is a well-posed function and $(\mathbf{W}^{(l)},\boldsymbol{b}^{(l)}) \in SP(\phi)$, we affirm that $\phi$ is Lipschitz continuous with constant $L_{\phi}$. Now, we start the mathematical induction. When $t=1$, for any $\boldsymbol{x}, \boldsymbol{x}' \in \mathbb{R}^{n_0}$, we have $\mathbb{E} \left[~ \| \boldsymbol{s}_t(\boldsymbol{x}) - \boldsymbol{s}_t(\boldsymbol{x}') \|_{\alpha}^{\textrm{sup}} ~\right] \leq C_{\eta,\theta,\alpha} \| \boldsymbol{x} - \boldsymbol{x}' \|_{\alpha}$, where $C_{\eta,\theta,\alpha} = \eta^{\alpha}~ \mathbb{E}[ |\mathcal{N}(0,1)|^{\alpha} ] $. Per mathematical induction, for $t \geq 1$, we have $\mathbb{E} \left[~ \| \boldsymbol{s}_t(\boldsymbol{x}) - \boldsymbol{s}_t(\boldsymbol{x}') \|_{\alpha}^{\textrm{sup}} ~\right] \leq C_{\eta,\theta,\alpha} \| \boldsymbol{x} - \boldsymbol{x}' \|_{\alpha} $. Thus, one has
\begin{equation} \label{eq:induction}
	\mathbb{E} \left[~ \| \boldsymbol{s}_t(\boldsymbol{x}) - \boldsymbol{s}_t(\boldsymbol{x}') \|_{\alpha}^{\textrm{sup}} ~\right] 
	\leq \frac{ (C_{\phi})^{\alpha} }{n_l} ~\mathbb{E}[~ |\mathcal{N}(0,1)|^{\alpha} ~]~ \Big\| \boldsymbol{s}_{t-1}(\boldsymbol{x}) - \boldsymbol{s}_{t-1}(\boldsymbol{x}') \Big\|_{\alpha}  \ ,
\end{equation}
where
\[
\begin{aligned}
	C_{\phi} &= \sigma^2_0(\boldsymbol{x}) - 2 \Sigma_{\boldsymbol{x},\boldsymbol{x}'} + \sigma^2_0(\boldsymbol{x}') \\
	&=  \frac{\eta^2}{n_l}~  \Big\| \phi\circ \boldsymbol{s}_{t-1}(\boldsymbol{x}) - \phi\circ \boldsymbol{s}_{t-1}(\boldsymbol{x}') \Big\|_2  \qquad\text{(~from Eq.~\eqref{eq:sigma}~)} \\
	&\leq \frac{\eta^2 L_{\phi}^2}{n_l}~ \big\| \boldsymbol{s}_{t-1}(\boldsymbol{x}) - \boldsymbol{s}_{t-1}(\boldsymbol{x}') \big\|_2 \ .
\end{aligned}
\]
Thus, Eq.~\eqref{eq:induction} becomes $\mathbb{E} [ \| \boldsymbol{s}_t(\boldsymbol{x}) - \boldsymbol{s}_t(\boldsymbol{x}') \|_{\alpha}^{\textrm{sup}} ]
\leq C'_{\eta,\theta,\alpha} \| \boldsymbol{x} - \boldsymbol{x}' \|_{\alpha}$, where $C_{\eta,\theta,\alpha}' = n_l^{-1}(\eta L_{\phi})^{\alpha} \| \boldsymbol{s}_{t-1}(\boldsymbol{x}) - \boldsymbol{s}_{t-1}(\boldsymbol{x}') \|_{\alpha} \mathbb{E}[~ |\mathcal{N}(0,1)|^{\alpha} ]$. Iterating this argument, we obtain $\mathbb{E} [ \| \boldsymbol{s}_t(\boldsymbol{x}) - \boldsymbol{s}_t(\boldsymbol{x}') \|_{\alpha}^{\textrm{sup}} ]  \leq C_{\eta,\theta,\alpha} \| \boldsymbol{x} - \boldsymbol{x}' \|_{\alpha}$, where $C_{\eta,\theta,\alpha} = \eta^{\alpha (t+1)} L_{\phi}^{\alpha t} ~ \mathbb{E}[~ |\mathcal{N}(0,1)|^{\alpha} ~]^{t+1}$. The above induction holds for any positive and even $\alpha$. Let $\beta = \alpha - n_0 > 0$, then this lemma is proved as desired. $\hfill\square$

\bibliography{JMref}

@inproceedings{yang2019:GP,
	title={Tensor programs {I}: Wide feedforward or recurrent neural networks of any architecture are Gaussian processes},
	author={G. Yang},
	booktitle={Advances in Neural Information Processing Systems 32},
	pages={9951--9960},
	year={2019}
}

@article{neal1996:GP,
	title={Priors for infinite networks},
	author={R. M. Neal},
	journal={Bayesian Learning for Neural Networks},
	pages={29--53},
	year={1996}
}

@inproceedings{cho2009:GP,
	title={Kernel methods for deep learning},
	author={Y. Cho and L. Saul},
	booktitle={Advances in Neural Information Processing Systems 22},
	pages={342--350},
	year={2009}
}

@inproceedings{novak2018:GP,
	title={Bayesian Deep Convolutional Networks with Many Channels are Gaussian Processes},
	author={R. Novak and L. Xiao and Y. Bahri and J. Lee and G. Yang and J. Hron and D. A. Abolafia and J. Pennington and J. Sohl-dickstein},
	booktitle={Proceedings of the 6th International Conference on Learning Representations},
	year={2018}
}

@inproceedings{garriga2019:GP,
	title={Deep Convolutional Networks as shallow Gaussian Processes},
	author={A. Garriga-Alonso and C. Rasmussen and L. Aitchison},
	booktitle={Proceedings of the 7th International Conference on Learning Representations},
	year={2019}
}

@inproceedings{lee2018:NNGP,
	title={Deep Neural Networks as Gaussian Processes},
	author={J. Lee and Y. Bahri and R. Novak and S. S. Schoenholz and J. Pennington and J. Sohl-Dickstein},
	booktitle={Proceedings of the 6th International Conference on Learning Representations},
	year={2018}
}

@article{pang2019:NNGP,
	title={Neural-net-induced Gaussian process regression for function approximation and {PDE} solution},
	author={G. Pang and L. Yang and G. E. Karniadakis},
	journal={Journal of Computational Physics},
	volume={384},
	pages={270--288},
	year={2019}
}

@article{park2020:NNGP,
	title={Towards {NNGP}-guided neural architecture search},
	author={D. S. Park and J. Lee and D. Peng and Y. Cao and J. Sohl-Dickstein},
	journal={arXiv:2011.06006},
	year={2020}
}

@article{zhang2022:nngp,
	title={Neural Network Gaussian Processes by Increasing Depth},
	author={S.-Q. Zhang and F. Wang and F.-L. Fan},
	journal={IEEE Transactions on Neural Networks and Learning Systems},
	volume={35},
	number={2},
	pages={2881--2886},
	year={2024}
}

@inproceedings{pleiss2022:NNGP,
	title={The limitations of large width in neural networks: A deep Gaussian process perspective},
	author={G. Pleiss and J. P. Cunningham},
	booktitle={Advances in Neural Information Processing Systems 34},
	pages={3349--3363},
	year={2021}
}

@inproceedings{jacot2018:NTK,
	title={Neural tangent kernel: Convergence and generalization in neural networks},
	author={A. Jacot and F. Gabriel and C. Hongler},
	booktitle={Advances in Neural Information Processing Systems 31},
	pages={8580 -- 8589},
	year={2018}
}

@inproceedings{arora2019:NTK,
	title={Fine-grained analysis of optimization and generalization for overparameterized two-layer neural networks},
	author={S. Arora and S. S. Du and W. Hu and Z. Li and R. Wang},
	booktitle={Proceedings of the 36th International Conference on Machine Learning},
	pages={322--332},
	year={2019}
}

@inproceedings{du2019:GNTK,
	title={Graph neural tangent kernel: Fusing graph neural networks with graph kernels},
	author={S. S. Du and K. Hou and R. R. Salakhutdinov and B. Poczos and R. Wang and K. Xu},
	booktitle={Advances in Neural Information Processing Systems 32},
	pages={5723 -- 5733},
	year={2019}
}

@inproceedings{huang2021:NTK,
	title={{FL-NTK}: A neural tangent kernel-based framework for federated learning analysis},
	author={B. Huang and X. Li and Z. Song and X. Yang},
	booktitle={Proceedings of the 38th International Conference on Machine Learning},
	pages={4423--4434},
	year={2021}
}

@article{mahankali2023:NTK,
	title={Beyond {NTK} with vanilla gradient descent: A mean-field analysis of neural networks with polynomial width, samples, and time},
	author={A. Mahankali and J. Z. Haochen and K. Dong and M. Glasgow and T. Ma},
	journal={arXiv:2306.16361},
	year={2023}
}

@inproceedings{malladi2023:NTK,
	title={A kernel-based view of language model fine-tuning},
	author={S. Malladi and A. Wettig and D. Yu and D. Chen and S. Arora},
	booktitle={Proceedings of the 40th International Conference on Machine Learning},
	pages={23610--23641},
	year={2023}
}

@inproceedings{hron2020:attention,
	title={Infinite attention: {NNGP} and {NTK} for deep attention networks},
	author={J. Hron and Y. Bahri and J. Sohl-Dickstein and R. Novak},
	booktitle={Proceedings of the 37th International Conference on Machine Learning},
	pages={4376--4386},
	year={2020}
}

@article{avidan2023:connecting,
	title={Connecting {NTK} and {NNGP}: A Unified Theoretical Framework for Neural Network Learning Dynamics in the Kernel Regime},
	author={Y. Avidan and Q. Li and H. Sompolinsky},
	journal={arXiv:2309.04522},
	year={2023}
}

@book{mezard1987:Replica,
	title={Spin glass theory and beyond: An Introduction to the Replica Method and Its Applications},
	author={M. M{\'e}zard and G. Parisi and M. A. Virasoro},
	year={1987},
	publisher={World Scientific Publishing Company}
}

@inproceedings{arora2019:NNGP,
	title={On exact computation with an infinitely wide neural net},
	author={S. Arora and S. S. Du and W. hu and Z. Li and R. R. Salakhutdinov and R. Wang},
	booktitle={Advances in Neural Information Processing Systems 32},
	pages={8141--8150},
	year={2019}
}

@inproceedings{bracale2020:asymptotic,
	title={Large-width functional asymptotics for deep gaussian neural networks},
	author={D. Bracale and S. Favaro and S. Fortini and S. Peluchetti},
	booktitle={Proceedings of the 8th International Conference on Learning Representations},
	year={2020}
}

@book{stroock1997:Kolmogorov,
	title={Multidimensional Diffusion Processes},
	author={D. Stroock and S. Varadhan},
	year={1997},
	publisher={Springer Science \& Business Media}
}

@article{zhang2021:arise,
	title={ARISE: ApeRIodic SEmi-parametric Process for Efficient Markets without Periodogram and Gaussianity Assumptions},
	author={S.-Q. Zhang and Z.-H. Zhou},
	journal={arXiv:2111.06222},
	year={2021}
}

@inproceedings{nguyen2021:eigenvalues,
	title={Tight bounds on the smallest eigenvalue of the neural tangent kernel for deep relu networks},
	author={Q. Nguyen and M. Mondelli and G. Montufar},
	booktitle={Proceedings of the 38th International Conference on Machine Learning},
	pages={8119--8129},
	year={2021}
}

@article{salas1999gershgorin,
	title={Gershgorin's theorem for matrices of operators},
	author={Salas, Hector N},
	journal={Linear Algebra and its Applications},
	volume={291},
	number={1-3},
	pages={15--36},
	year={1999}
}

@inproceedings{lee2020finite,
	title={Finite versus infinite neural networks: An empirical study},
	author={J. Lee and S. Schoenholz and J. Pennington and B. Adlam and L. Xiao and R. Novak and J. Sohl-Dickstein},
	booktitle={Advances in Neural Information Processing Systems 33},
	pages={15156--15172},
	year={2020}
}

@article{poggio2020theoretical,
	title={Theoretical issues in deep networks},
	author={T. Poggio and A. Banburski and Q. Liao},
	journal={Proceedings of the National Academy of Sciences},
	volume={117},
	number={48},
	pages={30039--30045},
	year={2020}
}

@book{van2000asymptotic,
	title={Asymptotic Statistics},
	author={Van der Vaart, A. W.},
	year={2000},
	publisher={Cambridge University Press}
}

@book{billingsley2013convergence,
	title={Convergence of Probability Measures},
	author={P. Billingsley},
	year={2013},
	publisher={John Wiley \& Sons}
}

@article{gou2021knowledge,
	title={Knowledge distillation: A survey},
	author={J. Gou and B. Yu and S. J. Maybank and D. Tao},
	journal={International Journal of Computer Vision},
	volume={129},
	number={6},
	pages={1789--1819},
	year={2021}
}

@article{zhuang2020:transfer,
	title={A comprehensive survey on transfer learning},
	author={F. Zhuang and Z. Qi and K. Duan and D. Xi and Y. Zhu and H. Zhu and H. Xiong and Q. He},
	journal={Proceedings of the IEEE},
	volume={109},
	number={1},
	pages={43--76},
	year={2020}
}

@article{hospedales2021:meta,
	title={Meta-learning in neural networks: A survey},
	author={T. Hospedales and A. Antoniou and P. Micaelli and A. Storkey},
	journal={IEEE Transactions on Pattern Analysis and Machine Intelligence},
	volume={44},
	number={9},
	pages={5149--5169},
	year={2021}
}
\bibliographystyle{plain}

\end{document}